\documentclass[letterpaper]{article} 
\usepackage{aaai2027}  
\usepackage[hyphens]{url}  
\usepackage{graphicx} 
\usepackage{natbib}  
\usepackage{caption} 
\usepackage{algorithm}
\usepackage{algorithmic}
\usepackage{newfloat}
\usepackage{listings}
\DeclareCaptionStyle{ruled}{labelfont=normalfont,labelsep=colon,strut=off} 
\floatstyle{ruled}
\newfloat{listing}{tb}{lst}{}
\floatname{listing}{Listing}

\usepackage{xcolor}
\usepackage[
    colorlinks=true,   
    citecolor=RoyalBlue,
    linkcolor=red,
    urlcolor=magenta
]{hyperref}

\setcitestyle{numbers,square,sort&compress}

\usepackage{booktabs}  
\usepackage{multirow}  
\usepackage{makecell}  
\usepackage[table,dvipsnames]{xcolor} 

\usepackage{pifont}
\newcommand{\cmark}{\textcolor{green!60!black}{\ding{52}}}
\newcommand{\xmark}{\textcolor{red}{\ding{56}}}

\usepackage{subcaption}
\usepackage{amsmath} 
\usepackage{enumitem}
\usepackage{fontawesome5} 

\usepackage{graphicx}
\usepackage{xspace}

\newcommand{\huggingface}{\raisebox{-1.5pt}{\includegraphics[height=1.05em]{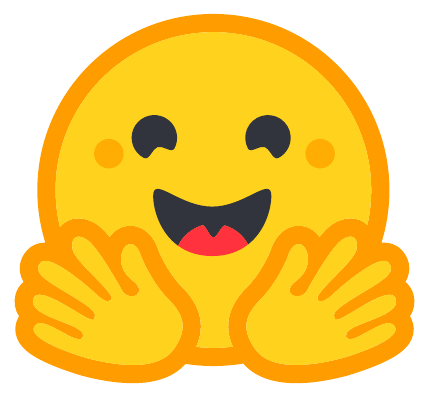}}\xspace}
\newcommand{\github}{\raisebox{-1.5pt}{\includegraphics[height=1.05em]{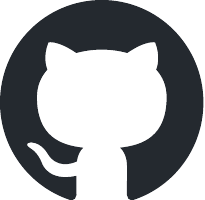}}\xspace}
\usepackage{cuted} 
\usepackage{fontawesome5}

\AtBeginDocument{
  \setcitestyle{numbers,square,comma,sort&compress}
}

\title{Unlocking the Potential of Image Editing via Concept Scaling and \\ Dense Supervision}
\author{
    Long Cui\textsuperscript{\rm 1,2},
    Xiaoqian Liu\textsuperscript{\rm 1},
    Qi Qin\textsuperscript{\rm 2},
    Yi Xin\textsuperscript{\rm 2},
    Tao Lin\textsuperscript{\rm 2},\\
    Jianguo Li\textsuperscript{\rm 2,\textdagger},
    Linfeng Zhang\textsuperscript{\rm 1,\textdagger}
}
\affiliations{
    \textsuperscript{\rm 1}Shanghai Jiao Tong University\\
    \textsuperscript{\rm 2}Ant Group
}

\nocopyright 

\begin{document}

\maketitle

\let\thefootnote\relax\footnotetext{\textsuperscript{\textdagger}Corresponding authors.}

\setlength{\stripsep}{-20pt} 

\begin{strip}
    \centering
    \renewcommand{\arraystretch}{1.2}
    \begin{tabular}{rll}
        \github{} & \textbf{GitHub Repo} & \url{https://github.com/inclusionAI/ConceptEdit} \\
        \huggingface{} & \textbf{HuggingFace Dataset} & \url{https://huggingface.co/collections/inclusionAI/conceptedit} \\
    \end{tabular}
    \vspace{4em}
\end{strip}

\begin{abstract}

Existing image editing frameworks predominantly follow the training paradigm of text-to-image diffusion models. However, extending this paradigm to image editing highlights two inherent discrepancies, specifically, the insufficient attention to edit concept granularity and the training inefficiency caused by sparse supervision signals. To address these issues, we establish a comprehensive hierarchical taxonomy featuring over 1,000 fine-grained edit concepts and build ConceptEdit-12M, a massive dataset of 12 million high-quality editing pairs via an improved synthesis framework. This library-driven approach effectively rectifies the distribution collapse of generated data while ensuring high data fidelity. Furthermore, we propose a dense supervision training strategy that synthesizes multiple non-interfering concepts into single image pairs. By providing richer learning signals, this strategy significantly enhances both training efficiency and overall model performance. Training results validate our strategy, significantly outperforming prior works. Finally, we present ConceptEdit-Bench, a granular evaluation suite designed to diagnose model capabilities across a vast array of real-world scenarios.

\end{abstract}
\section{Introduction}
\label{sec:intro}

Recent advancements in Text-to-Image (T2I) diffusion models \cite{rombach2022high, wu2025qwen, qin2025lumina, cai2025z, team2025longcat, google2025nanobanana, openai2025gpt4oimage} have achieved unprecedented success in generating diverse images of high quality. Building upon this, instruction-based image editing \cite{shi2024seededit, labs2025flux1kontextflowmatching, liu2025step1x-edit, wu2025qwen, dimoo, google2025nanobanana} has emerged as a crucial application, allowing users to modify specific aspects of an image while preserving its original context. Most existing editing frameworks adopt a training paradigm similar to T2I, where the model is conditioned on both the source image and textual instructions to predict the edited target. 

\begin{figure*}[t]      
    \centering            
    \includegraphics[width=1.0\textwidth]{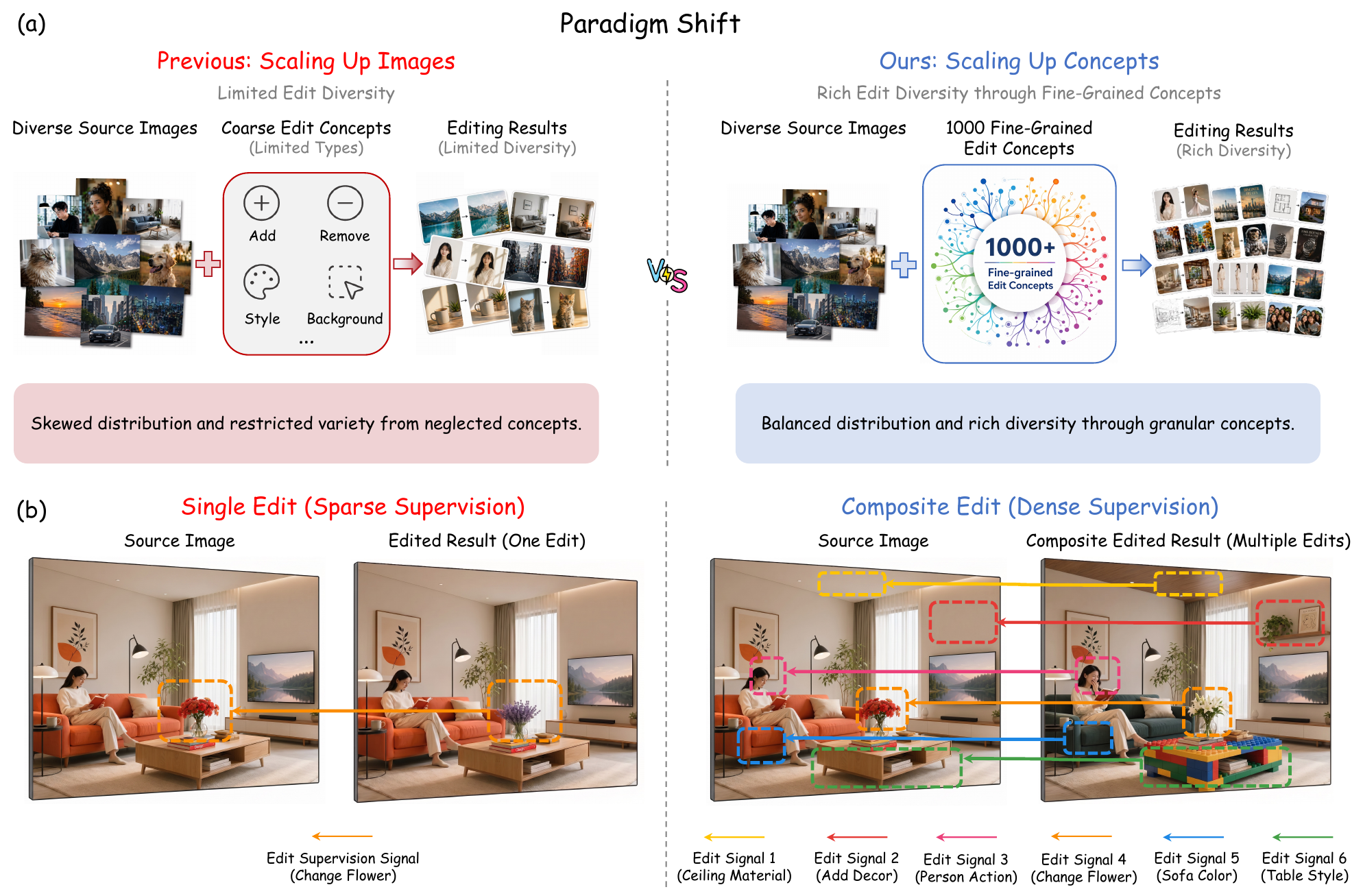} 
    \caption{
    \textbf{(a) Edit Concept Scaling.} Left: The previous paradigm is restricted by coarse categories and limited diversity. Right: Our approach scales up to 1,000+ fine-grained concepts to ensure a balanced and rich distribution. 
    \textbf{(b) Dense Supervision.} Left: Conventional training relies on single edit pairs with sparse supervision signals. Right: Our composite edit strategy provides dense supervision, enhancing training efficiency.
    }
    \label{fig:motivaton} 
\end{figure*}

However, directly translating this training paradigm to image editing reveals two fundamental discrepancies. First, in the realm of T2I generation, it is a broad consensus that scaling up the diverse distribution of training data \cite{openai2023dalle3, esser2024scalingrectifiedflowtransformers, podell2023sdxlimprovinglatentdiffusion} is the key to enhancing generation capabilities. In contrast, an image editing sample fundamentally consists of two components: the source image and the \textbf{edit concept} (e.g.,~adding objects, replacing backgrounds, and altering attributes). While existing datasets \cite{wei2024omniedit, yu2024anyedit, zhao2024ultraedit, ye2025imgedit} have primarily scaled through the expansion of source image diversity, they pay insufficient attention to the diversity and granularity of the edit concepts (Fig.~\ref{fig:motivaton}a). Second, unlike the global synthesis in T2I where generative signals apply to every pixel, image editing is fundamentally sparse. Specifically, only localized regions are actively modified, while the majority of the image serves as a static consistency constraint (Fig.~\ref{fig:motivaton}b). Consequently, current research overlooks both the granularity of edit concepts and the training inefficiency caused by such sparse supervision.

    
    
    
    

\begin{figure*}[t]      
    \centering            
    \includegraphics[width=1.0\textwidth]{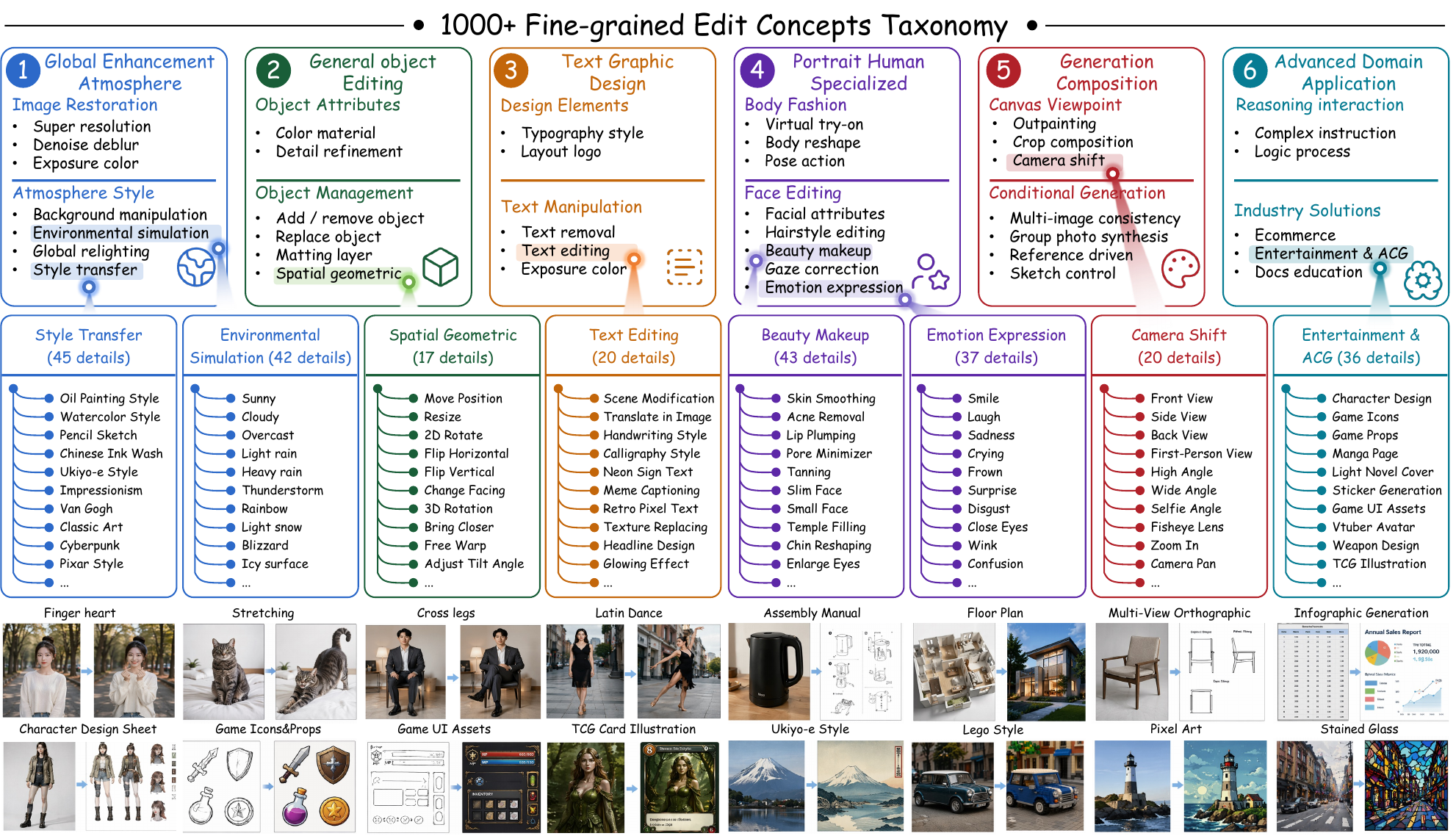} 
    \caption{\textbf{Hierarchical taxonomy of 1,000+ fine-grained edit concepts.} 
    Top: Overall hierarchical framework. 
    Mid: Specific leaf nodes for detailed edit concepts.
    Bottom: Visualizations of edit samples.}
    \label{fig:taxo} 
\end{figure*}

In response to the first observation, we argue that the diversity of edit concepts is the true bottleneck for generalization in image editing. In this work, we move beyond simple data scaling and systematically investigate the impact of edit concept granularity on model performance. We hypothesize that the primary bottleneck in image editing training arises not from a lack of source image variety, but from insufficient exposure to a fine-grained distribution of potential modifications. To explore this, we establish a comprehensive taxonomy across multiple levels that defines over 1,000 fine-grained edit concepts, aiming to exhaustively cover the vast landscape of editing scenarios in the real world. As illustrated in Fig.~\ref{fig:taxo}, this system subdivides coarse categories into precise modifications. For instance, character actions are refined into specific movements like ``finger heart'' or ``shrugging,'' while expressions are expanded to include nuanced states such as ``anxious'' or ``confused.'' Our empirical findings demonstrate that scaling the diversity of edit concepts more effectively unlocks the model's potential and enhances its overall editing capabilities. Furthermore, this taxonomy enables us to establish a comprehensive benchmark that evaluates editing models at a fine granularity.

To resolve the second discrepancy, we rethink the efficiency of editing supervision signals. While compositional editing has been introduced as a specific task in several studies \cite{ye2025imgedit, ye2025unicedit}, it is mostly treated as an end goal rather than a fundamental mechanism for training. We argue that the inherent sparsity of single edit samples, where often only a fraction of pixels provide active learning signals, limits training efficiency. However, we observe that localized edits are often distributed across spatial regions that do not interfere with one another, making them ideal candidates for compositional compression. To leverage this, we propose a training strategy that synthesizes multiple fine-grained edit concepts into a single image pair, thereby providing dense supervision signals. Our experiments reveal that training on these densely supervised samples not only enhances the model's capability for complex editing but also broadly improves its performance on ordinary single edits, providing a more efficient training paradigm for general image editing.

\begin{figure*}[t]      
    \centering            
    \includegraphics[width=0.99\textwidth]{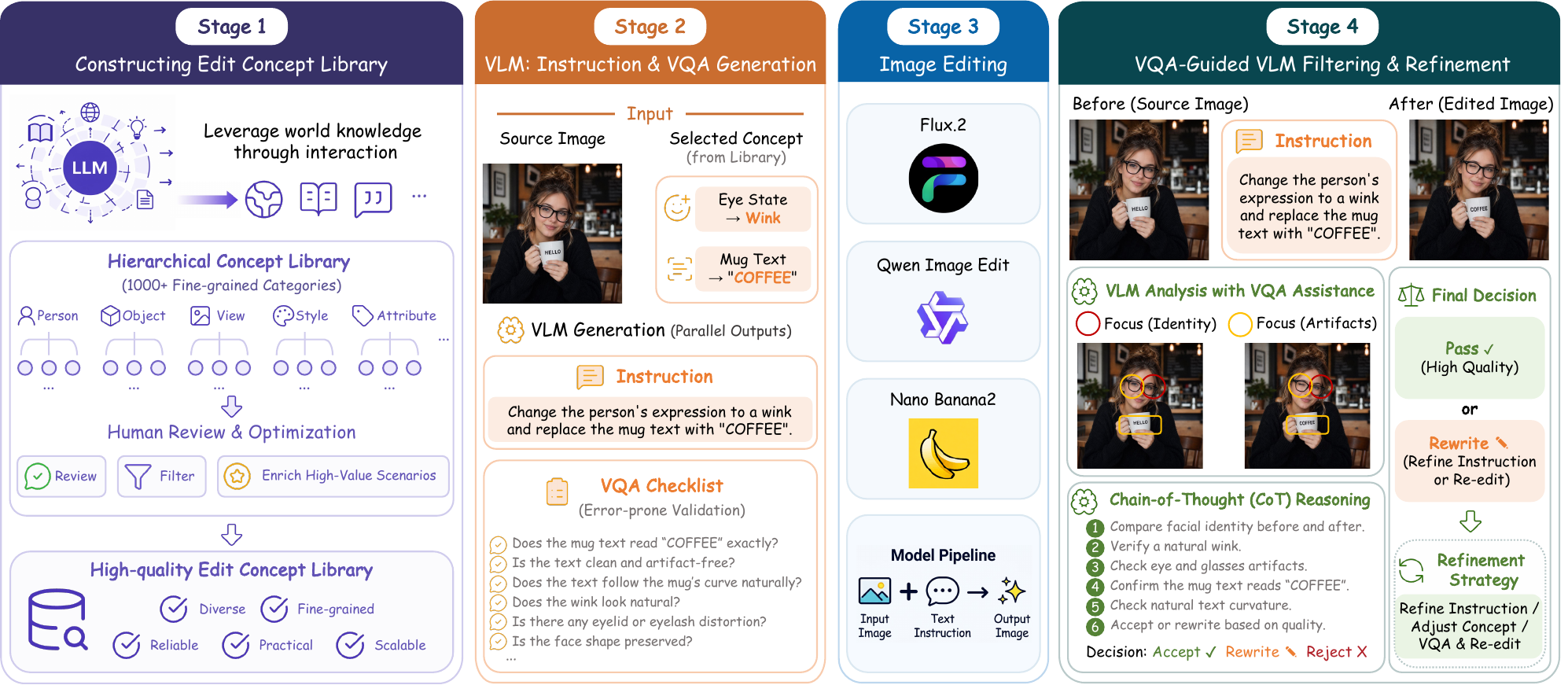} 
    \caption{\textbf{Overview of the improved synthesis framework.} 
    Stage 1: Library Construction leveraging LLM world knowledge. 
    Stage 2: Semantic Matching and Instruction Generation including VQA checklists. 
    Stage 3: Image Synthesis using various editing models. 
    Stage 4: Instance Specific Verification using VQA.}
    \label{fig:pipeline} 
\end{figure*}

To operationalize these insights and ensure high training data fidelity, we propose an improved synthesis framework (Fig.~\ref{fig:pipeline}) that enhances current editing pipelines~\cite{ye2025unicedit, chen2026scaleedit} by replacing stochastic sampling of coarse categories with a structured concept library. Rather than starting directly with instruction generation as in existing pipelines, we first distill extensive world knowledge from Large Language Models \cite{openai2023gpt4, google2024gemini} to proactively enumerate potential edit concepts across diverse domains. This allows us to controllably guide the distribution and diversity of the editing data to ensure exhaustive coverage of scenarios in the real world. To ensure the fidelity of our synthesis pipeline, we incorporate VQA filtering tailored to each instance. Rather than relying on generic templates, this mechanism generates customized pairs of questions and answers for every case. This guides the model to focus on localized regions prone to errors and directs a chain of thought (CoT) for systematic verification. This approach establishes a robust foundation for implementing concept scaling and dense supervision, supporting the construction of our dataset and a comprehensive, granular benchmark.

In summary, our main contributions are as follows:
\begin{itemize}[leftmargin=2.5em]
    \item \textbf{Edit Concept Scaling.} We propose a paradigm shift in edit data scaling, moving from source image variety to edit concept richness. We establish a hierarchical taxonomy of 1,000 fine-grained categories to investigate concept diversity and enhance training performance.
    
    \item \textbf{Dense Supervision Training.} We propose a training methodology that leverages compositional edits to provide dense supervision signals, boosting training efficiency and performance for tasks involving both single and multiple concepts.
    
    \item \textbf{Improved Synthesis Framework.} We develop a framework that distills LLM world knowledge for exhaustive scenario coverage. It incorporates VQA filtering tailored to each instance, directing focus toward localized regions and assisting CoT for verification.
    
    \item \textbf{ConceptEdit Dataset and Benchmark.} We construct a 12M high-quality editing dataset and benchmark across over 1,000 fine-grained categories, facilitating robust training and granular evaluation in real-world scenarios. To our knowledge, this scale represents the largest image editing dataset to date, tying with ScaleEdit-12M~\cite{chen2026scaleedit}.

\end{itemize}

\section{Related Work}
\label{sec:related_work}

\paragraph{Image Editing Models.}
Diffusion priors~\cite{rombach2022high, podell2023sdxl} catalyzed a paradigm shift in text-guided image editing. While early methods used manual latent engineering, InstructPix2Pix~\cite{brooks2023instructpix2pix} introduced instruction tuning, later refined by task-specific frameworks like OmniEdit~\cite{wei2024omniedit}. Concurrently, unified multimodal models, such as Bagel~\cite{deng2025bagel}, Emu3.5~\cite{cui2025emu3}, InternVL-U~\cite{tian2026internvl}, and LLaDA2.0-Uni~\cite{ai2026llada20uniunifyingmultimodalunderstanding}, demonstrate strong editing capabilities. However, open-source models still struggle to match the reasoning ceiling of proprietary systems like GPT-4o~\cite{openai2024gpt4o} and Nano Banana~\cite{google2025nanobanana}, highlighting the need for premium, knowledge-intensive datasets.

\paragraph{Image Editing Datasets.}
Image editing performance depends heavily on training data volume and diversity. Consequently, the field has transitioned from manual curation like MagicBrush~\cite{zhang2024magicbrushmanuallyannotateddataset} to automated pipelines. These pipelines synthesize data using multi-tool workflows in UltraEdit~\cite{zhao2024ultraedit}, ImgEdit~\cite{ye2025imgedit}, and Step1X-Edit~\cite{liu2025step1x-edit}, or generative models in NHR~\cite{kuprashevich2025no} and HQ-Edit~\cite{hui2024hqedit}. Recent efforts like UnicEdit~\cite{ye2025unicedit} and ScaleEdit~\cite{chen2026scaleedit} optimize these synthesis processes for refined data. However, despite growing data volume, existing datasets neglect edit concept distributions, limiting model generalization across real-world scenarios.

\section{Methodology}
\label{sec:method}



\begin{figure}[th] 
    \centering
    \begin{subfigure}[b]{0.49\columnwidth}
        \centering
        \includegraphics[width=\textwidth]{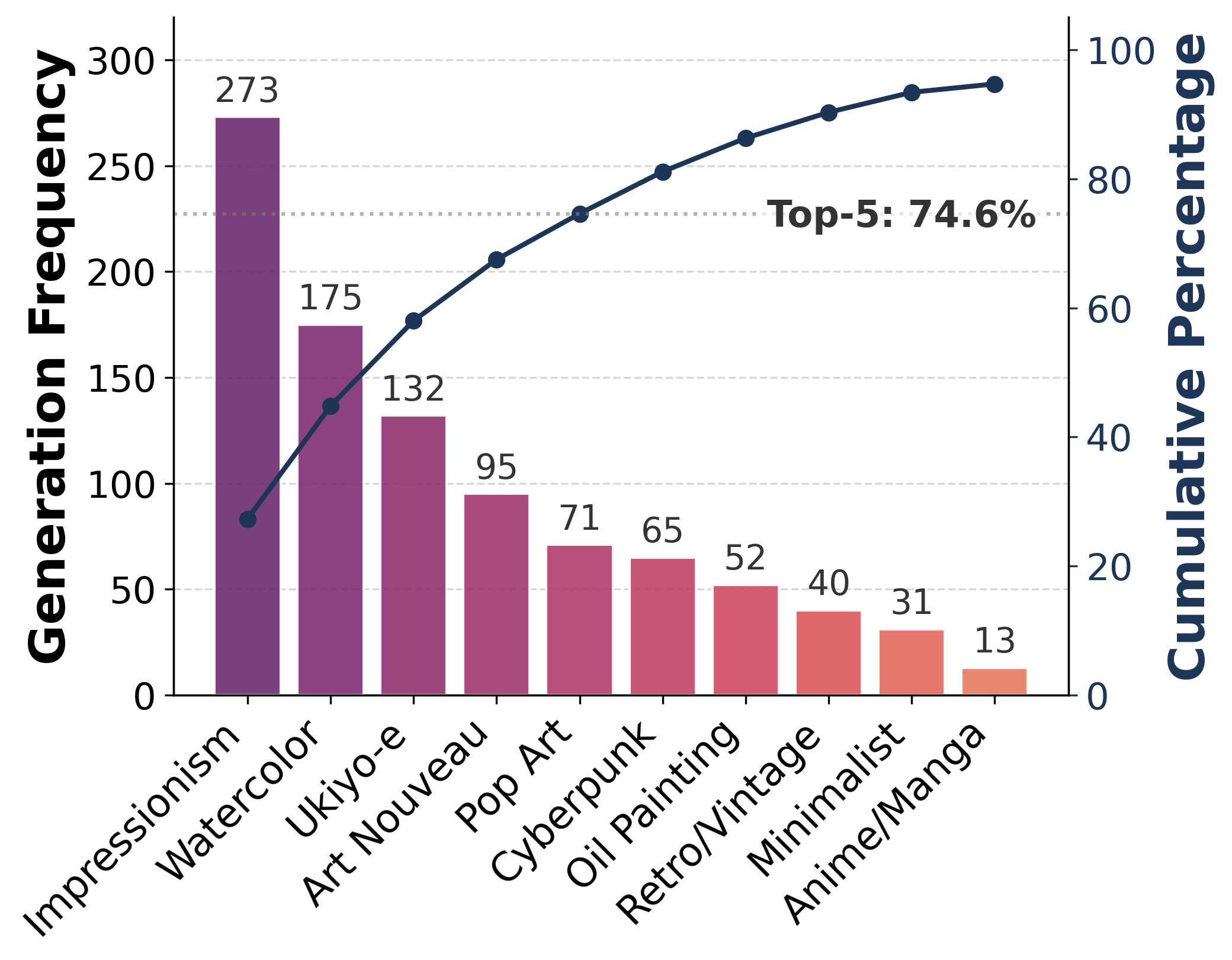}
        \caption{Stochastic Sampling}
        \label{fig:dist_vlm}
    \end{subfigure}
    \hfill
    \begin{subfigure}[b]{0.49\columnwidth}
        \centering
        \includegraphics[width=\textwidth]{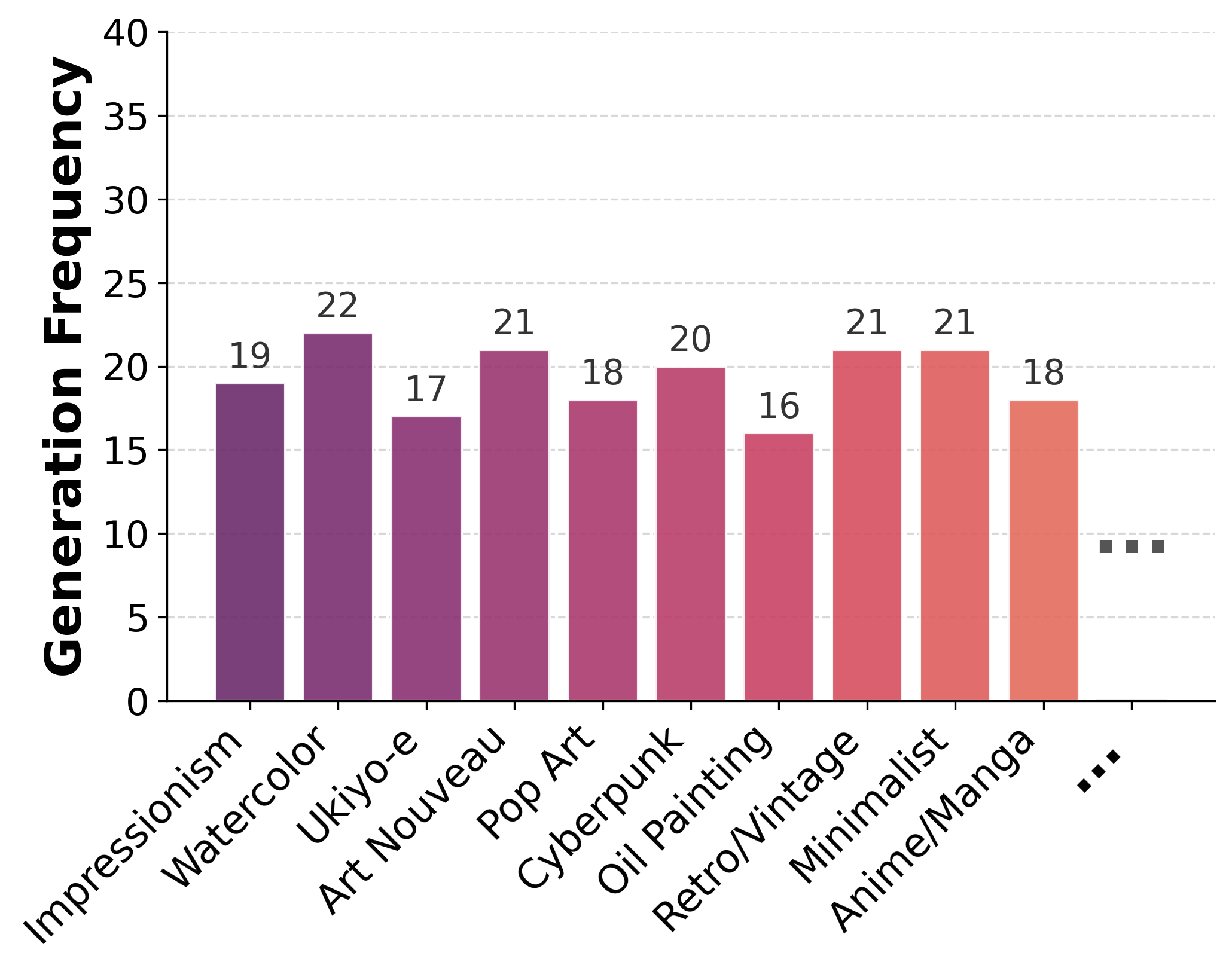}
        \caption{Balanced (Ours)}
        \label{fig:dist_ours}
    \end{subfigure}
    \caption{Edit concept distributions. Stochastic sampling collapses while our library ensures diversity.}
    \label{fig:distribution_comparison}
\end{figure}

\subsection{Rethinking Data: VLM Distribution Collapse}
\label{sec:rethinking_data}

Overcoming the generalization bottleneck in image editing requires scaling the diversity of edit concepts. However, current pipelines heavily rely on Vision-Language Models (VLMs) to stochastically generate instructions based on limited coarse-grained categories. This unconstrained dependency leads to a severe distribution collapse due to inherent VLM biases. As shown in Fig.~\ref{fig:distribution_comparison}, in the ``style transfer'' category, stochastic sampling causes the top 5 styles to dominate 74.6\% of the generated instructions, leaving dozens of others at less than 1\%. This collapse hinders domain generalization, where the ``domain'' is the precise edit concept defined by the instruction. Existing datasets thus draw sparse, biased samples from the vast instruction space. We therefore propose a paradigm shift from stochastic VLM generation to a structured, library-driven approach. Explicitly populating the instruction space with over 1,000 fine-grained categories ensures uniform exposure to diverse visual transformations and establishes a robust conceptual foundation.

\subsection{An Improved Synthesis Framework}
\label{subsec:improved_framework}

Based on the insights discussed in Sec.~\ref{sec:rethinking_data}, we propose an improved synthesis framework designed to generate high-fidelity image editing pairs with a controllable concept distribution. Improving upon existing frameworks \cite{ye2025unicedit, chen2026scaleedit}, our pipeline consists of four key stages, as shown in Fig.~\ref{fig:pipeline}: (1) \textbf{Library Construction}, where a structured Edit Concept Library is built to provide world knowledge; (2) \textbf{Semantic Matching and Instruction Generation}, where a VLM evaluates the compatibility between candidate concepts and specific source images to produce precise editing instructions and VQA verification metrics; (3) \textbf{Image Synthesis}, where an editing model is invoked to generate the target images based on the textual instructions; and (4) \textbf{Instance-Specific Verification}, where the results are filtered through localized VQA. Leveraging this improved pipeline, we produce \textbf{ConceptEdit-12M}, a large dataset containing 12 million verified, high-quality image editing pairs.

\subsubsection{Edit Concept Library Construction}
\label{subsubsec:concept_library}

The core of our framework lies in the construction of the comprehensive Edit Concept Library. While existing pipelines often rely on a handful of human-predefined coarse categories (typically 10--20), we aim to exhaustively cover the broad spectrum of common and valuable edit operations. As illustrated in Fig.~\ref{fig:taxo}, we scale these operations into over 1,000 fine-grained categories to ensure high conceptual density for robust generalization. To materialize this hierarchical taxonomy, we propose an automated, iterative workflow to distill world knowledge from Large Language Models. 

Specifically, starting with a lightweight, manually initialized seed taxonomy, the LLM is prompted to continuously evaluate and dynamically expand the classification tree. During each iteration, the LLM is tasked to: (1) merge or prune redundant concepts, (2) extrapolate new intermediate subcategories, and (3) populate highly specific leaf nodes across diverse domains (e.g., physical attributes, complex human actions). This self-expanding loop repeats until the semantic expansion converges, effectively exhausting the LLM's internal conceptual space for image manipulations. Finally, to ensure programmatic rigor, this LLM-generated taxonomy is meticulously refined by human experts to resolve remaining semantic overlaps and supplement missing high-value edge cases. As shown in Fig.~\ref{fig:taxo}, the resulting library comprises over 1,000 fine-grained edit concepts, providing a dense and structured conceptual space for downstream image editing.

\subsubsection{Semantic Matching and Instruction Generation}
\label{subsubsec:matching}

The extreme granularity of our library necessitates rigorous semantic grounding to ensure compatibility between concepts and image contexts (e.g., avoiding ``finger heart'' edits on landscapes). We employ a VLM generator $\Phi$. For a source image $\mathbf{I}$, we sample a candidate concept subset $\mathcal{C}_{\text{cand}} \subset \mathcal{C}_{\text{lib}}$ of size $N$, and formalize the matching process as:
\begin{equation}
\Phi \big( \mathbf{I}, \, \mathcal{C}_{\text{cand}} \big) \mapsto \big\{ (c_k, t_k, v_k) \big\}_{k=1}^M, \quad \text{s.t. } M \le N
\end{equation}
where $c_k \in \mathcal{C}_{\text{cand}}$ represents the matched concept, $t_k$ is the generated editing instruction, and $v_k$ denotes the  accompanying VQA verification criteria generated concurrently.
This fine-grained oversight enables dynamic distribution control. By tracking the frequencies of over 1,000 categories, we adaptively adjust sampling weights for $\mathcal{C}_{\text{cand}}$, which prevents distribution collapse and ensures conceptual diversity (Fig.~\ref{fig:dist_ours}). Furthermore, to avoid restriction by the static taxonomy, we introduce a stochastic exploration mechanism. With a predefined probability, the VLM can bypass $\mathcal{C}_{\text{cand}}$ to autonomously propose novel concepts from its world knowledge, making the dataset both structured and diverse.

\subsubsection{Instance-Specific VQA Filtering}
\label{subsubsec:vqa_filtering}

The quality of synthesized editing pairs depends on the precision of the filtering stage. VLM-based filtering typically utilizes generic prompts or predefined category-level templates for holistic assessment. However, these generalized approaches fail to account for the specific characteristics and potential failure modes of individual cases. To resolve this, we implement instance-specific VQA filtering as a supplementary verification layer.

Our framework utilizes the customized question-answer pairs $v_k$ generated during the instruction phase. In addition to the holistic assessment, these targeted queries direct the VLM to inspect localized regions that are particularly prone to editing failures. This structured inquiry facilitates CoT reasoning, enabling the VLM to systematically evaluate the correspondence between the instruction and the visual modification. By guiding the model to focus on critical, instruction-relevant areas while maintaining a global perspective, this mechanism improves the detection of subtle misalignments and provides a basis for either discarding low-quality samples or refining the associated instructions through recaptioning.

\begin{table*}[t]
\centering
\small 
\setlength{\tabcolsep}{6pt} 
\caption{\textbf{Quantitative comparison and ablation study on the ImgEdit benchmark~\cite{ye2025imgedit}.} The models are trained on 2M and 5M data scales respectively. Abbreviations: Ext.: Extract, Rm.: Remove, Bg.: Background, Adj.: Adjust, Rep.: Replace, Act.: Action, Comp.: Compose. $\Delta$ Comp. Gain indicates the improvement brought by dense supervision (w/ Comp) over ConceptEdit$_{1000}$. $\Delta$ Overall Gain highlights the total performance margin of our full framework over previous best-performing baseline (ScaleEdit). Best results per category at each scale are in bold.}
\label{tab:main_results}
\begin{tabular}{ll ccccccccc c}
\toprule
\multirow{2}{*}{\textbf{Scale}} & \multirow{2}{*}{\textbf{Method}} & \multicolumn{9}{c}{\textbf{Fine-Grained Editing Categories}} & \multirow{2}{*}{\textbf{Overall}} \\
\cmidrule(lr){3-11}
& & \textbf{Ext.} & \textbf{Add} & \textbf{Style} & \textbf{Rm.} & \textbf{Bg.} & \textbf{Adj.} & \textbf{Rep.} & \textbf{Act.} & \textbf{Comp.} & \\
\midrule
\multirow{8}{*}{2M} 
& UnicEdit           & 2.20 & 3.71 & 3.62 & 1.52 & 3.48 & 3.24 & 3.24 & 3.46 & 2.07 & 2.95 \\
& ScaleEdit         & 2.10 & 3.71 & 4.35 & \textbf{3.22} & 2.86 & 2.70 & \textbf{3.57} & 3.58 & 2.42 & 3.17 \\
\cmidrule(lr){2-12}
& ConceptEdit$_{10}$        & 2.18 & 3.82 & 3.61 & 2.00 & 3.51 & 3.27 & 2.69 & \textbf{3.83} & 2.53 & 3.05 \\
& ConceptEdit$_{500}$       & 2.19 & 4.02 & 4.49 & 2.46 & 3.58 & 3.66 & 2.43 & 3.78 & 2.75 & 3.26 \\
& ConceptEdit$_{1000}$    & 2.17 & 3.96 & 4.55 & 2.71 & 3.49 & 3.59 & 3.33 & 3.63 & 2.52 & 3.33 \\
& ConceptEdit$_{1000 \, \text{w/ Comp}}$ & \textbf{2.23} & \textbf{4.09} & \textbf{4.71} & 2.81 & \textbf{3.75} & \textbf{3.70} & 3.39 & 3.73 & \textbf{2.87} & \textbf{3.48} \\
\rowcolor[gray]{0.95}  & \quad \textit{$\Delta$ Comp. Gain} 
& \textcolor{RoyalBlue}{\textit{+0.06}} & \textcolor{RoyalBlue}{\textit{+0.13}} & \textcolor{RoyalBlue}{\textit{+0.16}} 
& \textcolor{RoyalBlue}{\textit{+0.10}} & \textcolor{RoyalBlue}{\textit{+0.26}} & \textcolor{RoyalBlue}{\textit{+0.11}} 
& \textcolor{RoyalBlue}{\textit{+0.06}} & \textcolor{RoyalBlue}{\textit{+0.10}} & \textcolor{RoyalBlue}{\textit{+0.35}} 
& \textcolor{RoyalBlue}{\textit{+0.15}} \\
\rowcolor[gray]{0.95}  & \textit{\textbf{$\Delta$ Overall Gain}} 
& \textcolor{RoyalBlue}{\textbf{\textit{+0.13}}} & \textcolor{RoyalBlue}{\textbf{\textit{+0.38}}} & \textcolor{RoyalBlue}{\textbf{\textit{+0.36}}} 
& \textcolor{gray}{\textit{-0.41}} & \textcolor{RoyalBlue}{\textbf{\textit{+0.89}}} & \textcolor{RoyalBlue}{\textbf{\textit{+1.00}}} 
& \textcolor{gray}{\textit{-0.18}} & \textcolor{RoyalBlue}{\textbf{\textit{+0.15}}} & \textcolor{RoyalBlue}{\textbf{\textit{+0.45}}} 
& \textcolor{RoyalBlue}{\textbf{\textit{+0.31}}} \\
\midrule
\multirow{8}{*}{5M} 
& UnicEdit           & 2.26 & 3.80 & 3.70 & 2.22 & 3.57 & 3.40 & 2.89 & 4.01 & 2.62 & 3.16 \\
& ScaleEdit         & 2.19 & 3.64 & 4.57 & \textbf{3.64} & 2.75 & 2.79 & \textbf{3.91} & 3.84 & 2.42 & 3.31 \\
\cmidrule(lr){2-12}
& ConceptEdit$_{10}$        & 2.15 & 3.97 & 4.66 & 2.01 & 3.66 & 3.60 & 3.04 & 3.70 & 2.64 & 3.27 \\
& ConceptEdit$_{500}$       & 2.35 & 4.05 & 4.75 & 2.56 & 3.82 & 3.59 & 3.12 & \textbf{4.30} & 2.75 & 3.48 \\
& ConceptEdit$_{1000}$    & 2.32 & 4.12 & \textbf{4.83} & 3.35 & \textbf{3.92} & 3.78 & 3.52 & 3.98 & 2.61 & 3.60 \\
& ConceptEdit$_{1000 \, \text{w/ Comp}}$ & \textbf{2.50} & \textbf{4.24} & 4.78 & 3.46 & 3.75 & \textbf{3.98} & 3.78 & 4.23 & \textbf{3.04} & \textbf{3.75} \\
\rowcolor[gray]{0.95}  & \quad \textit{$\Delta$ Comp. Gain} 
& \textcolor{RoyalBlue}{\textit{+0.18}} & \textcolor{RoyalBlue}{\textit{+0.12}} & \textcolor{gray}{\textit{-0.05}} 
& \textcolor{RoyalBlue}{\textit{+0.11}} & \textcolor{gray}{\textit{-0.17}} & \textcolor{RoyalBlue}{\textit{+0.20}} 
& \textcolor{RoyalBlue}{\textit{+0.26}} & \textcolor{RoyalBlue}{\textit{+0.25}} & \textcolor{RoyalBlue}{\textit{+0.43}} 
& \textcolor{RoyalBlue}{\textit{+0.15}} \\
\rowcolor[gray]{0.95}  & \textit{\textbf{$\Delta$ Overall Gain}} 
& \textcolor{RoyalBlue}{\textbf{\textit{+0.31}}} & \textcolor{RoyalBlue}{\textbf{\textit{+0.60}}} & \textcolor{RoyalBlue}{\textbf{\textit{+0.21}}} 
& \textcolor{gray}{\textit{-0.18}} & \textcolor{RoyalBlue}{\textbf{\textit{+1.00}}} & \textcolor{RoyalBlue}{\textbf{\textit{+1.19}}} 
& \textcolor{gray}{\textit{-0.13}} & \textcolor{RoyalBlue}{\textbf{\textit{+0.39}}} & \textcolor{RoyalBlue}{\textbf{\textit{+0.62}}} 
& \textcolor{RoyalBlue}{\textbf{\textit{+0.44}}} \\
\bottomrule
\end{tabular}
\end{table*}

\subsection{Dense Supervision via Composition}
\label{subsec:dense_supervision}

\begin{figure}[t] 
    \centering
    \includegraphics[width=0.65\columnwidth]{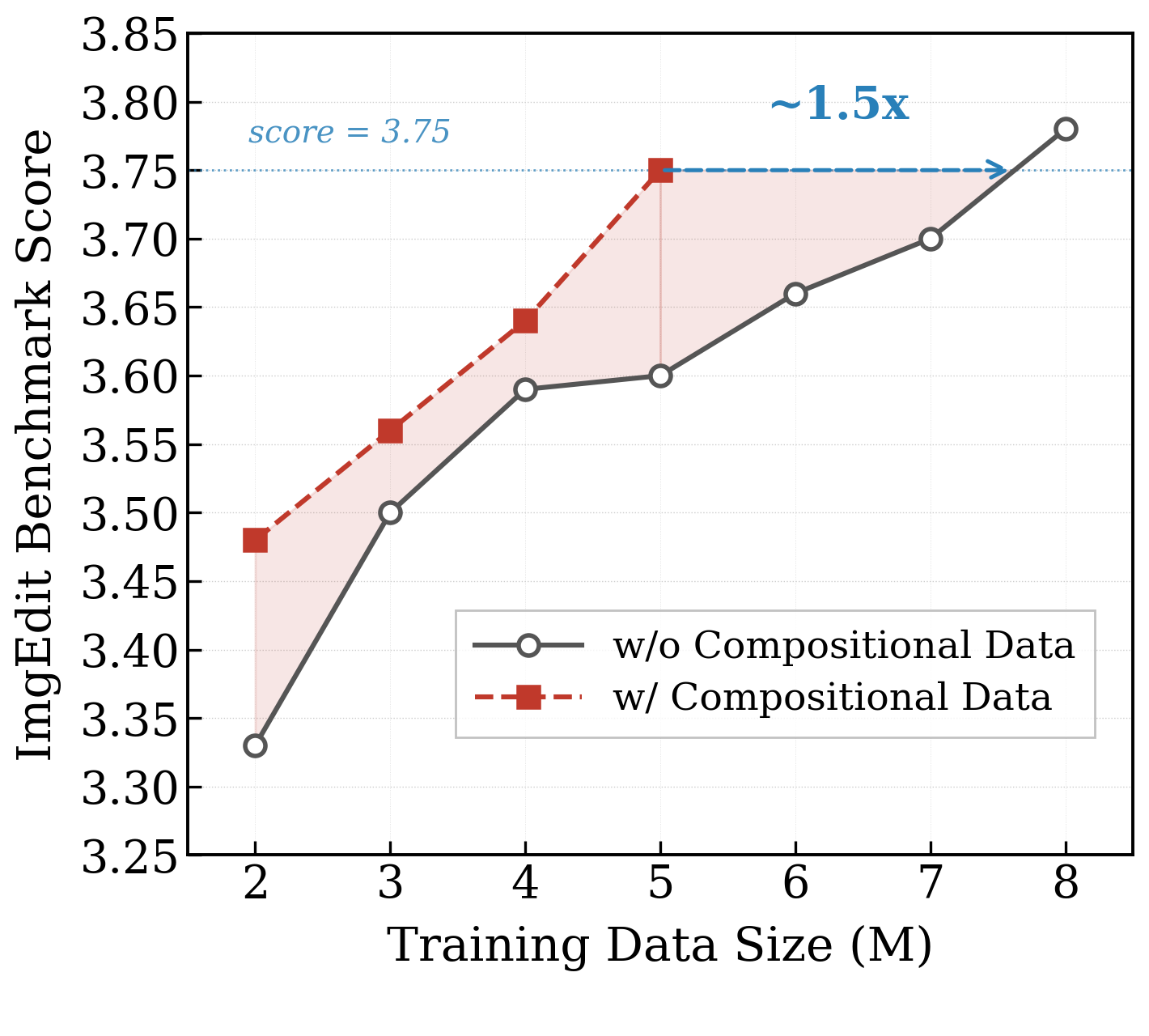}
    \caption{Training efficiency for different strategies.}
    \label{fig:efficiency_plot}
\end{figure}

Consistent with the observations in Sec.~\ref{sec:intro}, the inherent sparsity of single-concept edits limits overall training efficiency. As modified regions typically occupy only a small fraction of the image, the training objective becomes dominated by the reconstruction loss of the static background. This results in highly sparse editing supervision, where the model primarily optimizes for an identity mapping rather than active generative transformations.

To resolve this, we propose integrating multiple non-interfering edit concepts into a single image pair, a strategy we formulate as dense supervision via composition. To this end, we employ a VLM-driven aggregator $\Psi$ to perform compositional selection and instruction aggregation. For a source image $\mathbf{I}$, we sample a candidate concept subset $\{(c_n, m_n)\}_{n=1}^N$, where $c_n \in \mathcal{C}_{\text{lib}}$ is an edit concept and $m_n$ is its corresponding edit region, and formalize the process as:
\begin{equation}
\begin{aligned}
\Psi \big( \mathbf{I}, \, \{(c_n, m_n)\}_{n=1}^N \big)
&\mapsto \big( T_{\text{comp}}, V_{\text{comp}}, \{(c_k, m_k)\}_{k=1}^M \big), \\
\text{s.t. }&
m_i \cap m_j = \emptyset,\quad i\neq j, \quad M\le N .
\end{aligned}
\end{equation}
where $M$ denotes the number of successfully selected concepts. In this formulation, the constraint $m_i \cap m_j = \emptyset$ is enforced on the output set to ensure that the selected edit regions are spatially disjoint, effectively preventing visual or conceptual interference. The mapping produces a single unified instruction $T_{\text{comp}}$, a global verification checklist $V_{\text{comp}}$, and the set of selected concept pairs $\{(c_k, m_k)\}_{k=1}^M$.

This approach strategically distributes edit points across disparate regions to achieve balanced spatial coverage, while strictly mitigating regional overlaps to prevent visual or conceptual interference. Such complex data synthesis is facilitated by the high-fidelity framework detailed in Sec.~\ref{subsec:improved_framework}, where we perform single or multiple model invocations to sequentially or concurrently apply various modifications.
Our instance-specific VQA filtering serves as a linchpin in this process. By leveraging customized queries $V_{\text{comp}}$, the system can rigorously verify the execution of each independent edit within the composite pair. This strategy functions as a form of spatial data compression, significantly increasing the information entropy per sample. By providing dense supervision signals within a single forward pass, the model is forced to allocate more representation capacity to learning structural transformations rather than background preservation, empirically accelerating convergence (Fig.~\ref{fig:efficiency_plot}) and enhancing performance for both single and multi concept editing tasks.

\subsection{The ConceptEdit Benchmark}
\label{sec:benchmark}

While existing benchmarks for instruction-based image editing, such as ImgEdit-Bench~\cite{ye2025imgedit} and GEdit-Bench~\cite{liu2025step1x-edit}, have advanced the field, they focus on coarse-grained capabilities under 50 generic types. However, in practice, a high-level aggregate score often obfuscates critical model failures in complex or long-tail cases. For instance, a model may excel at general action changes but fail at precise gestures like a ``finger heart.'' To address this, ConceptEdit-Bench provides a ``microscopic'' view through over 1,000 fine-grained concepts, offering high controllability and modular diagnostic capability. Unlike benchmarks providing only a single aggregated score, our taxonomy allows selective monitoring of specific clusters of interest. This granular feedback is critical for iterative model development, enabling developers to precisely identify improvements in specific capabilities after training updates.

Leveraging our fine-grained concept library, we introduce ConceptEdit-Bench to test the limits of instruction-following precision. We selected 1,000 distinct editing categories from our library, ensuring that each represents a unique, fine-grained operation, such as distinguishing among a ``smile,'' ``smirk,'' and ``laugh.'' To guarantee broad visual distribution and high fidelity, source images are sampled from high-quality open-source datasets~\cite{kuznetsova2020openimages, ma2026finet2iopenlargescalediverse}, covering diverse categories. Benchmark results are provided in the Supplementary Appendix.

\section{Experiments}
\label{sec:experiments}

\begin{table*}[th]
\centering
\small 
\setlength{\tabcolsep}{13pt} 
\caption{\textbf{Quantitative comparison and ablation study on GEdit-Bench~\cite{liu2025step1x-edit}.} The models are evaluated at 2M and 5M training data scales. $\Delta$ Comp. Gain indicates the improvement brought by dense supervision (w/ Comp) over the baseline ConceptEdit$_{1000}$. $\Delta$ Overall Gain highlights the performance margin of our full framework over the previous best baseline (ScaleEdit). Best results per metric at each scale are in bold.}
\label{tab:gedit_bench_results}
\begin{tabular}{ll ccc ccc}
\toprule
\multirow{2}{*}{\textbf{Scale}} & \multirow{2}{*}{\textbf{Method}} & \multicolumn{3}{c}{\textbf{GEdit-Bench-EN}} & \multicolumn{3}{c}{\textbf{GEdit-Bench-CN}} \\
\cmidrule(lr){3-5} \cmidrule(lr){6-8}
& & \textbf{$G_{SC} \uparrow$} & \textbf{$G_{PQ} \uparrow$} & \textbf{$G_{O} \uparrow$} & \textbf{$G_{SC} \uparrow$} & \textbf{$G_{PQ} \uparrow$} & \textbf{$G_{O} \uparrow$} \\
\midrule
\multirow{8}{*}{2M} 
& UnicEdit           & 4.87 & 7.03 & 4.79 & 4.83 & 7.00 & 4.65 \\
& ScaleEdit         & 5.30 & 6.79 & 5.38 & 5.33 & 6.66 & 5.25 \\
\cmidrule(lr){2-8}
& ConceptEdit$_{10}$        & 5.34 & 6.76 & 5.41 & 5.54 & \textbf{7.03} & 5.15 \\
& ConceptEdit$_{500}$       & 5.90 & \textbf{7.13} & 5.54 & 5.51 & 6.70 & 5.45 \\
& ConceptEdit$_{1000}$    & 5.91 & 6.76 & 5.48 & 5.87 & 7.00 & 5.48 \\
& ConceptEdit$_{1000 \, \text{w/ Comp}}$ & \textbf{6.34} & 6.79 & \textbf{5.81} & \textbf{6.32} & 6.93 & \textbf{5.83} \\
\rowcolor[gray]{0.95}  & \quad \textit{$\Delta$ Comp. Gain} 
& \textcolor{RoyalBlue}{\textit{+0.43}} & \textcolor{RoyalBlue}{\textit{+0.03}} & \textcolor{RoyalBlue}{\textit{+0.33}} 
& \textcolor{RoyalBlue}{\textit{+0.45}} & \textcolor{gray}{\textit{-0.07}} & \textcolor{RoyalBlue}{\textit{+0.35}} \\
\rowcolor[gray]{0.95}  & \textit{\textbf{$\Delta$ Overall Gain}} 
& \textcolor{RoyalBlue}{\textbf{\textit{+1.04}}} & \textcolor{RoyalBlue}{\textbf{\textit{+0.00}}} & \textcolor{RoyalBlue}{\textbf{\textit{+0.43}}} 
& \textcolor{RoyalBlue}{\textbf{\textit{+0.99}}} & \textcolor{RoyalBlue}{\textbf{\textit{+0.27}}} & \textcolor{RoyalBlue}{\textbf{\textit{+0.58}}} \\
\midrule
\multirow{8}{*}{5M} 
& UnicEdit           & 5.39 & 7.17 & 5.25 & 5.32 & 7.13 & 5.07 \\
& ScaleEdit         & 5.77 & 6.69 & 5.77 & 5.62 & 6.97 & 5.63 \\
\cmidrule(lr){2-8}
& ConceptEdit$_{10}$        & 5.91 & 6.93 & 5.93 & 5.83 & 7.00 & 5.81 \\
& ConceptEdit$_{500}$       & 6.84 & 7.00 & 6.30 & 6.80 & 7.05 & 6.31 \\
& ConceptEdit$_{1000}$    & 6.86 & 7.19 & 6.40 & 6.75 & 7.31 & 6.36 \\
& ConceptEdit$_{1000 \, \text{w/ Comp}}$ & \textbf{7.07} & \textbf{7.30} & \textbf{6.62} & \textbf{7.07} & \textbf{7.42} & \textbf{6.60} \\
\rowcolor[gray]{0.95}  & \quad \textit{$\Delta$ Comp. Gain} 
& \textcolor{RoyalBlue}{\textit{+0.21}} & \textcolor{RoyalBlue}{\textit{+0.11}} & \textcolor{RoyalBlue}{\textit{+0.22}} 
& \textcolor{RoyalBlue}{\textit{+0.32}} & \textcolor{RoyalBlue}{\textit{+0.11}} & \textcolor{RoyalBlue}{\textit{+0.24}} \\
\rowcolor[gray]{0.95}  & \textit{\textbf{$\Delta$ Overall Gain}} 
& \textcolor{RoyalBlue}{\textbf{\textit{+1.30}}} & \textcolor{RoyalBlue}{\textbf{\textit{+0.61}}} & \textcolor{RoyalBlue}{\textbf{\textit{+0.85}}} 
& \textcolor{RoyalBlue}{\textbf{\textit{+1.45}}} & \textcolor{RoyalBlue}{\textbf{\textit{+0.45}}} & \textcolor{RoyalBlue}{\textbf{\textit{+0.97}}} \\
\bottomrule
\end{tabular}
\end{table*}

\subsection{Implementation Details}
\label{subsec:setup}
We conduct our experiments using the Z-Image~\cite{cai2025z} framework as our base model. We evaluate our method on ImgEdit-Bench and GEdit-Bench using training scales of 2M and 5M samples. To avoid confounding variables in ablations, all samples are synthesized using Qwen3.5-122B-A10B \cite{qwen2026qwen35} (instructions/filtering) and FLUX.2-klein-9B~\cite{bfl2026flux2klein} (image synthesis). We employ a constant learning rate of $1 \times 10^{-5}$ and a total batch size of 512. Other hyperparameters remain fixed for fair comparison, unless otherwise specified.

\subsection{Comparative Study}
\label{subsec:comparison}
We evaluate ConceptEdit against UnicEdit and ScaleEdit, two advanced open-source datasets, the latter of which has established its superiority through standardized training evaluations. All models are assessed on ImgEdit-Bench~\cite{ye2025imgedit} and GEdit-Bench~\cite{liu2025step1x-edit} at 2M and 5M scales across diverse categories to measure overall editing capability. For the 5M scale, UnicEdit utilized repeated samples as only a portion of its data has been released.

As reported in Table~\ref{tab:main_results}, $\text{ConceptEdit}_{1000 \, \text{w/ Comp}}$ consistently yields the highest overall scores on ImgEdit-Bench. At 2M and 5M scales, ConceptEdit achieves overall scores of 3.48 and 3.75, outperforming ScaleEdit by absolute margins of 0.31 and 0.44 points, respectively. ConceptEdit shows clear advantages in categories such as \textbf{Add}, \textbf{Style}, \textbf{Bg.}, and \textbf{Act.}. Similarly, results on GEdit-Bench (Table~\ref{tab:gedit_bench_results}) demonstrate the consistent superiority of ConceptEdit across both English and Chinese evaluations. Notably, the performance boost is primarily attributed to the increased accuracy in instruction following ($G_{SC}$), which aligns with our theoretical expectations. These findings underscore the efficacy of scaling edit concepts to enhance model generalization.
\subsection{Ablation Study}
\label{subsec:ablation}

\subsubsection{Effect of Concept Scaling}
\label{subsubsec:scaling}
To investigate the impact of edit concept diversity on model performance, we compare three variants of our dataset with increasing granularity: ConceptEdit$_{10}$, ConceptEdit$_{500}$, and ConceptEdit$_{1000}$. 
On ImgEdit-Bench, scaling from 10 to 500 and 1,000+ categories improves 2M-scale scores from 3.05 to 3.26 and 3.33, respectively. At the 5M scale, ConceptEdit$_{1000}$ (3.60) outperforms ConceptEdit$_{10}$ (3.27) by 0.33 points. GEdit-Bench corroborates this trend. At the 5M scale, moving from 10 to 500 concepts boosts $G_{SC}$ for English (5.91 to 6.84) and Chinese (5.83 to 6.80) evaluations, with ConceptEdit$_{1000}$ maintaining these levels. These gains span tasks like \textbf{Style}, \textbf{Adj.}, and \textbf{Rep.}, proving that fine-grained concept distribution effectively outperforms naive scaling with coarse instructions.

\subsubsection{Effect of Dense Supervision}
\label{subsubsec:dense}
To evaluate dense supervision, we mix ConceptEdit$_{1000}$ with composite edits in a 1:1 ratio. 
As shown in Table~\ref{tab:main_results}, this composition consistently improves ImgEdit-Bench scores by 0.15 points across scales. These gains extend beyond the \textbf{Comp.} task, boosting categories like \textbf{Adj.} (+0.20), \textbf{Rep.} (+0.26), and \textbf{Act.} (+0.25) at the 5M scale. On GEdit-Bench-EN, the $G_{\text{SC}}$ score at the 2M scale corroborates this with a \textit{$\Delta$ Comp. Gain} of up to 0.43 points, confirming that learning from spatially non-interfering composite edits improves core visual understanding. Furthermore, matching the \textit{w/ Comp} performance without composite data requires $1.5\times$ more samples (Fig.~\ref{fig:efficiency_plot}). This validates that compositional edits provide dense supervision, significantly compressing training time.

\subsubsection{Effect of VQA Filtering}
\label{subsubsec:vqa}
\begin{table}[th]
\centering
\caption{Ablation study of our VQA filtering pipeline.}
\label{tab:filtering_ablation}
\small 
\setlength{\tabcolsep}{2pt} 
\begin{tabular}{l cccc} 
\toprule
\multirow{2}{*}{\textbf{Filtering Strategy}} & \textbf{Precision} & \textbf{Recall} & \textbf{F1-Score} & \textbf{Accuracy} \\
& \textbf{(\%)} & \textbf{(\%)} & \textbf{(\%)} & \textbf{(\%)} \\
\midrule
Pseudo GT (Gemini)  & 100.0 & 100.0 & 100.0 & 100.0 \\
\midrule
Generic Validation     & 75.0  & 57.0  & 65.0  & 90.0  \\
Instance-Specific (Ours) & 84.0  & 87.0  & 86.0  & 95.0  \\
\textit{\textbf{$\Delta$ Gain}} & 
\textbf{\textit{$\uparrow$ 9.0}} & 
\textbf{\textit{$\uparrow$ 30.0}} & 
\textbf{\textit{$\uparrow$ 21.0}} & 
\textbf{\textit{$\uparrow$ 5.0}} \\
\bottomrule
\end{tabular}

\end{table}
To ensure high fidelity alignment and suppress hallucinations, we compare our instance-specific VQA filtering against a generic VLM validation baseline. While the baseline uses uniform prompts for holistic assessment, the ConceptEdit pipeline dynamically formulates fine-grained questions tailored to specific edit concepts (e.g., checking for localized artifacts). Metrics are evaluated against pseudo-labels from Gemini-3-Pro~\cite{googlegemini3pro}.
As shown in Table~\ref{tab:filtering_ablation}, our VQA pipeline significantly outperforms the baseline, boosting Precision, Recall, F1-score, and Accuracy by 9.0\%, 30.0\%, 21.0\%, and 5.0\%, respectively. While generic validation often overlooks subtle failures due to salient object bias, our region-aware verification successfully detects localized artifacts. This customized approach proves highly effective at identifying hallucinations, ensuring high-quality training data.

\section{Conclusion}
\label{sec:conclusion}
This work addresses the lack of edit concept granularity and sparse training signals in instruction-based image editing. We introduce ConceptEdit, a structured paradigm emphasizing edit concept scaling and dense supervision. Specifically, we build a 1,000-category hierarchical taxonomy and leverage composite edits for training. Our experiments show that granular edit concepts significantly enhance editing capabilities across diverse scenarios. Additionally, dense supervision accelerates training convergence by $1.5\times$ and boosts performance on single-concept tasks. Our instance-specific VQA filtering also reduces errors compared to generic validation. Finally, we present the ConceptEdit-12M dataset and ConceptEdit-Bench. This suite achieves SOTA results, outperforming existing baselines like ScaleEdit and UnicEdit.

\bibliography{AnonymousSubmission2027} 

\clearpage
\appendix
\setcounter{page}{1}
\section*{Supplementary Material}

\section{Discussion on Generalized I2I Translation}
\label{sec:appendix_generalized_i2i}

Conceptually, any image-to-image (I2I) translation can be viewed as a generalized form of image editing, where the source image serves as a structural condition and the text instruction specifies the target domain mapping. 

Our taxonomy of 1,000+ concepts primarily targets daily, user-centric interactive editing. We do not exhaustively categorize highly specialized or structural translation tasks, as they are typically treated as professional rendering or conditional generation rather than common interactive edits.

Nonetheless, to ensure broad scenario coverage and evaluate our model's adaptability, we incorporate a representative subset of classic structural tasks into our dataset, including:
\begin{itemize}
    \item Canny edges~\cite{canny1986computational} 
    \item HED edges~\cite{xie15hed} 
    \item Hough lines~\cite{gu2021realtime} 
    \item Semantic segmentation maps~\cite{uniformer} 
    \item Depth maps~\cite{depth_anything_v2} 
    \item Shape normal maps~\cite{xu2023unifying} 
    \item Human keypoints~\cite{openpose} 
\end{itemize}
This integration demonstrates that our framework remains robust and compatible with traditional, structurally constrained image translation paradigms.

\begin{figure}[htbp]
    \centering
    \includegraphics[width=\linewidth]{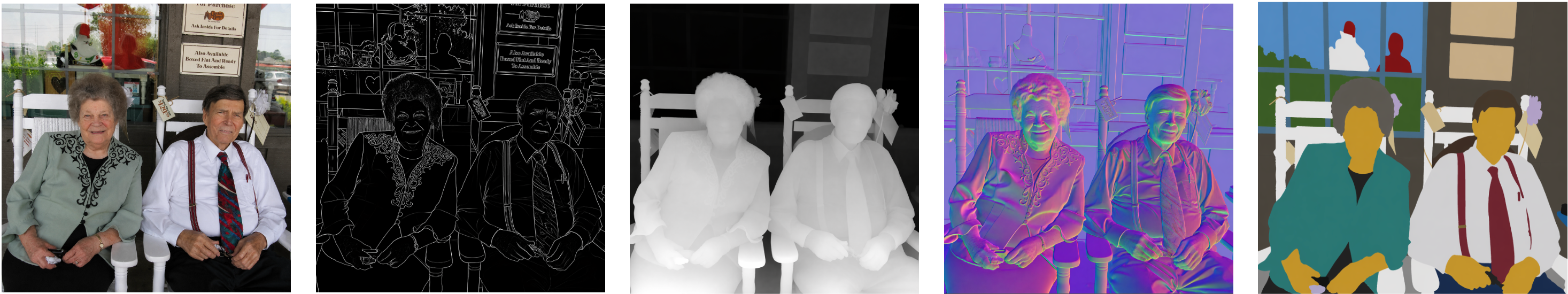}
    \caption{Visualizations of generalized image-to-image translation under various structural control signals.}
    \label{fig:condition_samples}
\end{figure}

As illustrated in Fig.~\ref{fig:condition_samples}, our dataset effectively supports these structurally conditioned transformations.

\section{Detailed Ablation on VQA Filtering Strategy}
\label{sec:appendix_vqa}

Table~\ref{tab:filtering_ablation_full} presents the complete ablation results of our VQA filtering pipeline, including the raw confusion matrix counts. 

By dynamically formulating tailored questions, our instance-specific strategy significantly reduces False Negatives (FN) from 72 to 21 and increases True Positives (TP) from 95 to 146 compared to the generic validation baseline. This targeted, region-aware verification effectively suppresses subtle edit failures, resulting in substantial gains across all metrics: $+9.0\%$ in Precision, $+30.0\%$ in Recall, $+21.0\%$ in F1-Score, and $+5.0\%$ in Accuracy.

\begin{table}[htbp]
\centering
\caption{Detailed ablation of our VQA filtering pipeline, including raw confusion matrix counts.}
\label{tab:filtering_ablation_full}
\small 
\setlength{\tabcolsep}{5pt}
\begin{tabular}{l ccc c}
\toprule
\textbf{Metric} & \textbf{Ground Truth} & \textbf{Generic} & \textbf{Ours} & \textbf{$\Delta$ Gain} \\
\midrule
TN             & 833   & 802  & 805  & -- \\
FN             & 0     & 72   & 21   & -- \\
TP             & 167   & 95   & 146  & -- \\
FP             & 0     & 31   & 28   & -- \\
\midrule
Precision (\%) & 100.0 & 75.0 & 84.0 & \textbf{\textit{$\uparrow$ 9.0}} \\
Recall (\%)    & 100.0 & 57.0 & 87.0 & \textbf{\textit{$\uparrow$ 30.0}} \\
F1-Score (\%)  & 100.0 & 65.0 & 86.0 & \textbf{\textit{$\uparrow$ 21.0}} \\
Accuracy (\%)  & 100.0 & 90.0 & 95.0 & \textbf{\textit{$\uparrow$ 5.0}} \\
\bottomrule
\end{tabular}

\end{table}

We also evaluate the per-sample computational overhead on Qwen3.5-122B-A10B. As shown in Table~\ref{tab:vqa_latency}, our instance-specific strategy incurs only a marginal total overhead of +0.069s per sample. Crucially, the entire data synthesis pipeline is heavily dominated by the DiT image generation phase, whereas the instruction generation and filtering stages together account for merely 2\%--10\% of the total runtime (depending on the image generation model used).

\begin{table}[htbp]
\centering
\caption{Per-sample processing latency (seconds) benchmarked on Qwen3.5-122B-A10B.}
\label{tab:vqa_latency}
\small
\setlength{\tabcolsep}{8pt}
\begin{tabular}{l cc c}
\toprule
\textbf{Stage} & \textbf{Generic} & \textbf{Ours} & \textbf{Overhead} \\
\midrule
Instruction Generation & 0.523s & 0.567s & +0.044s \\
Filtering Stage        & 0.312s & 0.337s & +0.025s \\
\midrule
\textbf{Total}          & 0.835s & 0.904s & +0.069s \\
\bottomrule
\end{tabular}

\end{table}

\section{Dataset Comparison}

As compared in Table~\ref{tab:comp_datasets}, we evaluate our proposed ConceptEdit dataset against existing mainstream image editing datasets across multiple dimensions. The comparison highlights several key advantages of ConceptEdit:
First, it significantly scales up both data volume and task diversity, containing 12 million editing pairs across over 1,000 subtasks, which far exceeds the maximum of 23 found in previous datasets.
Second, ConceptEdit establishes a more rigorous quality control pipeline. Notably, it is the only dataset that incorporates instance-specific filtering, which ensures high-fidelity alignment between editing instructions and visual changes. 
Third, it comprehensively supports both basic and complex editing scenarios, providing a more versatile resource for the community.
Lastly, ConceptEdit is the only dataset that features distribution control, allowing for a more balanced and controllable training.

\begin{table*}[ht]
\centering
\caption{Comparison of Image Editing datasets.}
\label{tab:comp_datasets}
\renewcommand{\arraystretch}{0.9}
\setlength{\tabcolsep}{2.5pt} 
\begin{tabular}{lcccccccc}
\toprule
\multirow{2}{*}{\textbf{Dataset}} & \multirow{2}{*}{\shortstack[c]{\textbf{Sub-}\\\textbf{tasks}}} & \multirow{2}{*}{\textbf{Size}} & \multicolumn{3}{c}{\textbf{Post Verification}} & \multirow{2}{*}{\shortstack[c]{\textbf{Basic}\\\textbf{Edit}}} & \multirow{2}{*}{\shortstack[c]{\textbf{Complex}\\\textbf{Edit}}} & \multirow{2}{*}{\shortstack[c]{\textbf{Dist.}\\\textbf{Control}}} \\ 
\cmidrule(lr){4-6} 
& & & \shortstack[c]{Failed\\Filtration} & \shortstack[c]{Inst.-Spec.\\Filtering} & Recaption & & & \\
\midrule\midrule 
MagicBrush~\cite{zhang2023magicbrush}            & 5   & $\sim$10K & \cmark & \xmark & \xmark & \cmark & \xmark & \xmark \\
InstructPix2Pix~\cite{brooks2023instructpix2pix} & 4   & $\sim$313K & \xmark & \xmark & \xmark & \cmark & \xmark & \xmark \\
HQ-Edit~\cite{hui2024hqedit}                     & 6   & $\sim$197K & \xmark & \xmark & \cmark & \cmark & \xmark & \xmark \\
SEED-Data-Edit~\cite{ge2024seeddataedit}         & 6   & $\sim$3.7M & \cmark & \xmark & \xmark & \cmark & \xmark & \xmark \\
UltraEdit~\cite{zhao2024ultraedit}               & 9   & $\sim$4M & \xmark & \xmark & \xmark & \cmark & \xmark & \xmark \\
ImgEdit~\cite{ye2025imgedit}                     & 13  & $\sim$1.2M & \cmark & \xmark & \xmark & \cmark & \cmark & \xmark \\
NHR-Edit~\cite{kuprashevich2025nohumansrequired} & -   & $\sim$358K & \cmark & \xmark & \xmark & \cmark & \xmark & \xmark \\
GPT-Image-Edit-1.5M~\cite{wang2025gptimageedit}  & -   & $\sim$1.5M & \xmark & \xmark & \cmark & \cmark & \xmark & \xmark \\
UnicEdit~\cite{ye2025unicedit}                   & 22  & $\sim$10M & \cmark & \xmark & \cmark & \cmark & \cmark & \xmark \\
ScaleEdit~\cite{chen2026scaleedit}               & 23  & $\sim$12M & \cmark & \xmark & \xmark & \cmark & \cmark & \xmark \\
\midrule
\textbf{ConceptEdit}                             & \textbf{1000} & \textbf{$\sim$12M} & \cmark & \cmark & \cmark & \cmark & \cmark & \cmark \\
\bottomrule
\end{tabular}

\end{table*}

\section{Model Performance on ConceptEdit-Bench} 
\begin{table*}[ht]
\centering
\caption{Quantitative comparison on ConceptEdit-Bench. The ``Overall'' column represents the final aggregated performance across our 1,000 fine-grained categories.}
\label{tab:conceptbench}
\begin{tabular}{lccccccc}
\toprule
\textbf{Model} & \textbf{Advanced} & \textbf{Object} & \textbf{Compo.} & \textbf{Global} & \textbf{Portrait} & \textbf{Text} & \textbf{Overall} \\
\midrule
\multicolumn{8}{c}{\textbf{Open-source Models}} \\
BAGEL-7B-MoT~\cite{deng2025bagel} & 42.21 & 46.12 & 38.85 & 44.90 & 41.10 & 33.91 & 41.66 \\
FLUX.1-Kontext-dev~\cite{labs2025flux1kontextflowmatching} & 50.73 & 42.88 & 44.45 & 49.05 & 37.36 & 40.15 & 43.53 \\
LongCat-Image-Edit~\cite{meituanlongcatteam2025longcatimagetechnicalreport} & 64.06 & 61.46 & 57.51 & 65.39 & 51.68 & 62.08 & 59.05 \\
Qwen-Image-Edit-2509~\cite{wu2025qwen} & 67.36 & 63.83 & 58.37 & 64.59 & 54.27 & 69.85 & 61.27 \\
Qwen-Image-Edit-2511~\cite{wu2025qwen} & 66.78 & 68.45 & 60.38 & 70.39 & 53.75 & 73.50 & 63.29 \\
FireRed-Image-Edit-1.0~\cite{superintelligenceteam2026fireredimageedit10technicalreport} & 70.12 & 72.40 & 63.52 & 71.21 & 56.42 & 74.72 & \textbf{65.86} \\
\midrule
\multicolumn{8}{c}{\textbf{Closed-source Models}} \\
Seedream 4.5~\cite{seedream2025seedream40nextgenerationmultimodal} & 69.10 & 68.00 & 63.62 & 68.60 & 52.90 & 71.32 & 63.34 \\
Nano Banana 2~\cite{google2025nanobanana} & 71.82 & 73.90 & 64.15 & 70.01 & 56.24 & 75.28 & \textbf{66.19} \\
\bottomrule
\end{tabular}

\end{table*}
Table~\ref{tab:conceptbench} presents a comprehensive evaluation of mainstream models on ConceptEdit-Bench. Closed-source models demonstrate strong performance, with Nano Banana 2 achieving the highest overall score of 66.19, while Seedream 4.5 records a score of 63.34. Within the open-source category, FireRed-Image-Edit-1.0 leads with a score of 65.86. A primary observation is that although most models successfully follow basic instructions, nearly all experience a notable decline in the Portrait and Composition categories. This indicates a general difficulty in performing edits that rely on detailed world knowledge, such as micro-expressions, or sophisticated spatial reasoning. Such findings highlight the value of our high-quality dataset and structured synthesis framework in addressing these persistent challenges.

\section{Semantic Overlap Across Categories}
\label{supp:concept_overlap}

In constructing our 1,000+ fine-grained edit taxonomy, we deliberately permit moderate semantic overlaps across different high-level categories rather than enforcing strict mutual exclusion. We consider that such concept overlap across categories is largely harmless to the overall dataset distribution and can inherently enhance instructional and visual diversity.

Specifically, we view a single visual attribute or physical primitive as carrying distinct operational semantics depending on its underlying intent and editing context. For instance, in our framework, the concept of ``neon lighting'' naturally manifests across multiple categories with subtle functional nuances: in Style Transfer, it serves as a holistic stylistic constraint governing global aesthetics; in Environmental Simulation, it acts as a plausible ambient atmospheric element; and in Global Relighting, it functions as an explicit light source manipulation that dictates surface reflections and shadow interactions.

In our view, strictly purging these intersections might impose artificial boundaries that decouple concepts from their real-world contexts. We hypothesize that retaining them allows the dataset to expose the model to identical visual primitives under varied instructional phrasing and generative goals, thereby mitigating rote pattern matching and encouraging greater model flexibility.

\section{Rationale for Downstream Evaluation}
\label{supp:evaluation_rationale}

While evaluating datasets via single-image aesthetic scores is an intuitive alternative, we deliberately prioritize downstream model training performance as our primary evaluation protocol. 

We consider that the core value of our dataset lies in its \textit{macro-level distribution richness, conceptual balance, and task coverage}, rather than merely maximizing the visual polish of isolated samples. A dataset containing aesthetically pleasing images can still suffer from severe distribution collapse if it lacks diverse edit operations. Because single-image aesthetic scores fail to measure conceptual diversity or instruction alignment, in our view, downstream training gains provide a far more rigorous indicator of dataset utility and real-world generalization.

\section{Hardware and Computing Infrastructure}
\label{supp:hardware}
All model training is conducted on NVIDIA H100 GPUs. Data synthesis, VQA filtering, and benchmark evaluations are performed on NVIDIA H20 GPUs. The entire framework is implemented in PyTorch and executed within a Linux environment.

\section{Dataset and Code Availability}
The ConceptEdit dataset, evaluation benchmark, and source code will be publicly released prior to or upon paper publication.

\section{Concept Library}
Our concept library is organized as a three-level taxonomy consisting of 1028 fine-grained edit concepts. The hierarchy below lists all leaf concepts in the library.

\begingroup
\small
\setlength{\parindent}{0pt}
\setlength{\parskip}{2pt}
\vspace{3pt}\noindent\rule{\linewidth}{0.4pt}\par
\noindent\textbf{1. Global Enhancement and Atmosphere}\par\vspace{1pt}
\noindent\hspace*{0.3em}\textbf{Image Restoration and Enhancement}\par
\noindent\hspace*{0.7em}\textbf{Super Resolution}\par
\noindent\hspace*{0.7em}\begin{tabular}{@{}p{0.45\linewidth}@{\hspace{0.02\linewidth}}p{0.45\linewidth}@{}}
$\bullet$ Image Upscaling & $\bullet$ Old Photo Restoration \\
$\bullet$ Anime Super-Resolution & $\bullet$ Text Sharpening \\
$\bullet$ Screenshot Repair & $\bullet$ Facial Detail Reconstruction \\
$\bullet$ Hair Detail Restoration & $\bullet$ Lossless Zoom \\
$\bullet$ Night Scene Clarity &  \\
\end{tabular}\par\vspace{1pt}
\noindent\hspace*{0.7em}\textbf{Denoise Deblur}\par
\noindent\hspace*{0.7em}\begin{tabular}{@{}p{0.45\linewidth}@{\hspace{0.02\linewidth}}p{0.45\linewidth}@{}}
$\bullet$ Motion Blur Removal & $\bullet$ Camera Shake Correction \\
$\bullet$ Out-of-Focus Repair & $\bullet$ Lens Blur Removal \\
$\bullet$ High ISO Denoising & $\bullet$ Film Grain Removal \\
$\bullet$ Moire Pattern Removal & $\bullet$ Dehaze \\
$\bullet$ Smart Sharpening & $\bullet$ Edge Enhancement \\
$\bullet$ Night Sight Clarity & $\bullet$ Screen Pattern Removal \\
$\bullet$ Face Deblur & $\bullet$ Background Denoising \\
\end{tabular}\par\vspace{1pt}
\noindent\hspace*{0.7em}\textbf{Exposure Color}\par
\noindent\hspace*{0.7em}\begin{tabular}{@{}p{0.45\linewidth}@{\hspace{0.02\linewidth}}p{0.45\linewidth}@{}}
$\bullet$ Low Light Enhancement & $\bullet$ Tone Unification \\
$\bullet$ Overexposure Repair & $\bullet$ Backlight Correction \\
$\bullet$ Dehaze & $\bullet$ Color Cast Removal \\
$\bullet$ HDR Effect & $\bullet$ Shadow \& Highlight Recovery \\
$\bullet$ Vibrance \& Saturation & $\bullet$ Contrast Enhancement \\
$\bullet$ Brightness Adjustment & $\bullet$ Color Match \\
$\bullet$ Faded Color Restoration & $\bullet$ Skin Tone Correction \\
$\bullet$ Cinematic Color Grading &  \\
\end{tabular}\par\vspace{1pt}
\noindent\hspace*{0.3em}\textbf{Atmosphere and Style}\par
\noindent\hspace*{0.7em}\textbf{Background Manipulation}\par
\noindent\hspace*{0.7em}\begin{tabular}{@{}p{0.45\linewidth}@{\hspace{0.02\linewidth}}p{0.45\linewidth}@{}}
$\bullet$ Remove background & $\bullet$ Transparent background \\
$\bullet$ Blur background & $\bullet$ Bokeh effect \\
$\bullet$ E-commerce white background & $\bullet$ Solid black background \\
$\bullet$ ID photo blue & $\bullet$ ID photo red \\
$\bullet$ Gradient background & $\bullet$ Morandi colors \\
$\bullet$ Desaturate background & $\bullet$ Darken background \\
$\bullet$ Sky replacement & $\bullet$ Product podium \\
$\bullet$ Studio lighting & $\bullet$ Marble texture \\
$\bullet$ Wooden tabletop & $\bullet$ Silk background \\
$\bullet$ Nature landscape & $\bullet$ City street \\
$\bullet$ Office interior & $\bullet$ Home interior \\
$\bullet$ Instagram style & $\bullet$ Cyberpunk background \\
$\bullet$ Minimalist geometry & $\bullet$ Holiday atmosphere \\
$\bullet$ Abstract art &  \\
\end{tabular}\par\vspace{1pt}
\noindent\hspace*{0.7em}\textbf{Environmental Simulation}\par
\noindent\hspace*{0.7em}\begin{tabular}{@{}p{0.45\linewidth}@{\hspace{0.02\linewidth}}p{0.45\linewidth}@{}}
$\bullet$ Sunny & $\bullet$ Cloudy \\
$\bullet$ Overcast & $\bullet$ Light rain \\
$\bullet$ Heavy rain & $\bullet$ Thunderstorm \\
$\bullet$ Rainbow & $\bullet$ Light snow \\
$\bullet$ Blizzard & $\bullet$ Snow accumulation \\
$\bullet$ Frost & $\bullet$ Icy surface \\
$\bullet$ Heavy fog & $\bullet$ Mist \\
$\bullet$ Haze & $\bullet$ Sandstorm \\
$\bullet$ Windy & $\bullet$ Sunrise \\
$\bullet$ Sunset & $\bullet$ Golden hour \\
$\bullet$ Blue hour & $\bullet$ Midnight \\
$\bullet$ Starry sky & $\bullet$ Moonlight \\
$\bullet$ Aurora & $\bullet$ God rays \\
$\bullet$ Lens flare & $\bullet$ Bokeh effects \\
$\bullet$ Fireflies & $\bullet$ Falling petals \\
$\bullet$ Floating dust & $\bullet$ Wet pavement \\
$\bullet$ Puddle reflections & $\bullet$ Spring bloom \\
$\bullet$ Summer vibe & $\bullet$ Autumn leaves \\
$\bullet$ Winter chill & $\bullet$ Cyberpunk neon \\
$\bullet$ Post-apocalyptic & $\bullet$ Dreamy atmosphere \\
$\bullet$ Gloomy atmosphere & $\bullet$ Underwater caustics \\
\end{tabular}\par\vspace{1pt}
\noindent\hspace*{0.7em}\textbf{Style Transfer}\par
\noindent\hspace*{0.7em}\begin{tabular}{@{}p{0.45\linewidth}@{\hspace{0.02\linewidth}}p{0.45\linewidth}@{}}
$\bullet$ Oil Painting Style & $\bullet$ Watercolor Style \\
$\bullet$ Pencil Sketch & $\bullet$ Chinese Ink Wash \\
$\bullet$ Ukiyo-e Style & $\bullet$ Impressionism/Van Gogh \\
$\bullet$ Classic Art/Renaissance & $\bullet$ Cyberpunk \\
$\bullet$ Steampunk & $\bullet$ Pixel Art \\
$\bullet$ 3D Cartoon/Pixar Style & $\bullet$ Ghibli/Anime Style \\
$\bullet$ American Comic Style & $\bullet$ Flat Illustration \\
$\bullet$ Low Poly & $\bullet$ Claymation \\
$\bullet$ Glitch Art & $\bullet$ Vaporwave \\
$\bullet$ Pop Art & $\bullet$ Graffiti/Street Art \\
$\bullet$ Paper Cut/Origami & $\bullet$ Stained Glass \\
$\bullet$ Mosaic Art & $\bullet$ Neon Noir \\
$\bullet$ Vintage Film/Retro & $\bullet$ Film Noir/High Contrast B\&W \\
$\bullet$ Crayon/Doodle Style & $\bullet$ Game CG/Concept Art \\
$\bullet$ Line Art & $\bullet$ Woodblock Print \\
$\bullet$ Charcoal Drawing & $\bullet$ Makoto Shinkai Style \\
$\bullet$ Lego/Block Style & $\bullet$ Relief/Emboss Style \\
$\bullet$ Gothic Dark Style & $\bullet$ Fantasy Fairy Tale \\
$\bullet$ Wasteland Style & $\bullet$ Rococo \\
$\bullet$ Surrealism & $\bullet$ Chalk Drawing \\
$\bullet$ Voxel Art & $\bullet$ Polaroid Style \\
$\bullet$ Acid Graphics & $\bullet$ Memphis Design \\
$\bullet$ Mechanical/Metallic Style &  \\
\end{tabular}\par\vspace{1pt}
\noindent\hspace*{0.7em}\textbf{Global Relighting}\par
\noindent\hspace*{0.7em}\begin{tabular}{@{}p{0.45\linewidth}@{\hspace{0.02\linewidth}}p{0.45\linewidth}@{}}
$\bullet$ Change Light Direction & $\bullet$ Golden Hour \\
$\bullet$ Blue Hour & $\bullet$ Sunset Glow \\
$\bullet$ Noon Sunlight & $\bullet$ Overcast Soft Light \\
$\bullet$ Moonlight & $\bullet$ Rembrandt Lighting \\
$\bullet$ Butterfly Lighting & $\bullet$ Rim Light \\
$\bullet$ Studio Soft Light & $\bullet$ Spotlight/Hard Light \\
$\bullet$ Window Light & $\bullet$ Neon Lighting \\
$\bullet$ Cinematic Lighting & $\bullet$ Volumetric Rays (God Rays) \\
$\bullet$ Candlelight Atmosphere & $\bullet$ Stage Lighting \\
$\bullet$ Underwater Caustics & $\bullet$ Face Fill Light \\
$\bullet$ Remove Shadows & $\bullet$ Cast Shadows \\
$\bullet$ Match Background Lighting & $\bullet$ Silhouette Effect \\
$\bullet$ Side Backlight &  \\
\end{tabular}\par\vspace{1pt}
\vspace{3pt}\noindent\rule{\linewidth}{0.4pt}\par
\noindent\textbf{2. General Object and Entity Editing}\par\vspace{1pt}
\noindent\hspace*{0.3em}\textbf{Object Management and Manipulation}\par
\noindent\hspace*{0.7em}\textbf{Add Remove Object}\par
\noindent\hspace*{0.7em}\begin{tabular}{@{}p{0.45\linewidth}@{\hspace{0.02\linewidth}}p{0.45\linewidth}@{}}
$\bullet$ Remove passersby & $\bullet$ Remove clutter \\
$\bullet$ Remove power lines & $\bullet$ Remove fences \\
$\bullet$ Remove vehicles & $\bullet$ Remove trash \\
$\bullet$ Remove street signs & $\bullet$ Remove glasses \\
$\bullet$ Remove jewelry & $\bullet$ Remove tattoos \\
$\bullet$ Remove reflections & $\bullet$ Remove shadows \\
$\bullet$ Add furniture & $\bullet$ Add plants \\
$\bullet$ Add decorations & $\bullet$ Add animals \\
$\bullet$ Add accessories & $\bullet$ Add props \\
$\bullet$ Generative fill & $\bullet$ Magic eraser \\
$\bullet$ Generate in area & $\bullet$ Universal Addition \\
$\bullet$ Universal Removal &  \\
\end{tabular}\par\vspace{1pt}
\noindent\hspace*{0.7em}\textbf{Replace Object}\par
\noindent\hspace*{0.7em}\begin{tabular}{@{}p{0.45\linewidth}@{\hspace{0.02\linewidth}}p{0.45\linewidth}@{}}
$\bullet$ Text-Guided Replacement & $\bullet$ Reference Image Replacement \\
$\bullet$ Keep Shape Replacement & $\bullet$ Free Form Replacement \\
$\bullet$ Swap Object Positions & $\bullet$ Generate Variations \\
$\bullet$ Replace Furniture & $\bullet$ Replace Decor \\
$\bullet$ Replace Plants \& Flowers & $\bullet$ Replace Vehicles \\
$\bullet$ Replace Handheld Objects & $\bullet$ Replace Wall Art \& Posters \\
$\bullet$ Replace Food \& Drinks & $\bullet$ Replace Electronics \\
$\bullet$ Replace Signage & $\bullet$ Replace Packaging \\
$\bullet$ Replace Background Props & $\bullet$ Replace Animals \\
$\bullet$ Replace Character Subject & $\bullet$ Replace Sculptures \\
$\bullet$ Replace Buildings & $\bullet$ Replace Ground Surface \\
$\bullet$ Replace Sky & $\bullet$ Universal Replacement and Modification \\
\end{tabular}\par\vspace{1pt}
\noindent\hspace*{0.7em}\textbf{Spatial Geometric}\par
\noindent\hspace*{0.7em}\begin{tabular}{@{}p{0.45\linewidth}@{\hspace{0.02\linewidth}}p{0.45\linewidth}@{}}
$\bullet$ Move Position & $\bullet$ Resize \\
$\bullet$ 2D Rotate & $\bullet$ Flip Horizontal \\
$\bullet$ Flip Vertical & $\bullet$ Change Object Facing \\
$\bullet$ 3D Object Rotation & $\bullet$ Perspective Correction \\
$\bullet$ Bring Object Closer & $\bullet$ Push Object Back \\
$\bullet$ Straighten Object & $\bullet$ Free Warp \\
$\bullet$ Mesh Transform & $\bullet$ Match Background Perspective \\
$\bullet$ Center Object & $\bullet$ Non-rigid Deformation \\
$\bullet$ Adjust Tilt Angle &  \\
\end{tabular}\par\vspace{1pt}
\noindent\hspace*{0.7em}\textbf{Matting Layer}\par
\noindent\hspace*{0.7em}\begin{tabular}{@{}p{0.45\linewidth}@{\hspace{0.02\linewidth}}p{0.45\linewidth}@{}}
$\bullet$ One-click Background Removal & $\bullet$ Make Background Transparent \\
$\bullet$ Portrait Matting & $\bullet$ Product Cutout \\
$\bullet$ Refine Hair Details & $\bullet$ Pet \& Animal Cutout \\
$\bullet$ Split Foreground \& Background & $\bullet$ ID Photo Cutout \\
$\bullet$ Extract Sky & $\bullet$ Extract Text or Logo \\
$\bullet$ Vehicle Cutout & $\bullet$ Clothing Segmentation \\
$\bullet$ Head/Face Cutout & $\bullet$ Smart Object Selection \\
$\bullet$ Generate Alpha Mask & $\bullet$ Edge Refinement \& Smoothing \\
$\bullet$ Green/Blue Screen Keying & $\bullet$ Food Cutout \\
$\bullet$ Complex Background Matting & $\bullet$ Batch Matting \\
\end{tabular}\par\vspace{1pt}
\noindent\hspace*{0.3em}\textbf{Object Attribute Refinement}\par
\noindent\hspace*{0.7em}\textbf{Color Material}\par
\noindent\hspace*{0.7em}\begin{tabular}{@{}p{0.45\linewidth}@{\hspace{0.02\linewidth}}p{0.45\linewidth}@{}}
$\bullet$ Precise local recoloring & $\bullet$ Smart object recoloring \\
$\bullet$ Change clothing fabric color & $\bullet$ Change vehicle color \\
$\bullet$ Product color variant generation & $\bullet$ Colorize black \& white photo \\
$\bullet$ Reference color transfer & $\bullet$ Turn into gold material \\
$\bullet$ Turn into silver chrome & $\bullet$ Turn into transparent glass \\
$\bullet$ Turn into jade gemstone & $\bullet$ Turn into marble texture \\
$\bullet$ Turn into solid wood & $\bullet$ Turn into ceramic glaze \\
$\bullet$ Turn into leather texture & $\bullet$ Turn into silk satin \\
$\bullet$ Turn into denim fabric & $\bullet$ Turn into plush fur \\
$\bullet$ Turn into knitted wool & $\bullet$ Turn into rusty metal \\
$\bullet$ Turn into neon glowing & $\bullet$ Turn into jelly gummy \\
$\bullet$ Turn into LEGO bricks & $\bullet$ Turn into origami paper \\
$\bullet$ Turn into clay plasticine & $\bullet$ Apply matte finish \\
$\bullet$ Apply glossy polish & $\bullet$ Add camouflage pattern \\
$\bullet$ Add floral pattern & $\bullet$ Add geometric plaid \\
$\bullet$ Material aging weathering &  \\
\end{tabular}\par\vspace{1pt}
\noindent\hspace*{0.7em}\textbf{Detail Refinement}\par
\noindent\hspace*{0.7em}\begin{tabular}{@{}p{0.45\linewidth}@{\hspace{0.02\linewidth}}p{0.45\linewidth}@{}}
$\bullet$ Remove wrinkles & $\bullet$ Remove scratches \\
$\bullet$ Remove stains & $\bullet$ Remove dust \\
$\bullet$ Remove fingerprints & $\bullet$ Remove glare \\
$\bullet$ Remove moire patterns & $\bullet$ Remove lint/pilling \\
$\bullet$ Repair cracks & $\bullet$ Repair damage \\
$\bullet$ Smooth edges & $\bullet$ Remove rust \\
$\bullet$ Remove mold & $\bullet$ Remove local shadows \\
$\bullet$ Enhance texture & $\bullet$ Repair peeling paint \\
$\bullet$ Remove sticker residue & $\bullet$ Leather repair \\
$\bullet$ Metal polishing & $\bullet$ Ceramic repair \\
$\bullet$ Glass repair & $\bullet$ Red-eye removal \\
\end{tabular}\par\vspace{1pt}
\vspace{3pt}\noindent\rule{\linewidth}{0.4pt}\par
\noindent\textbf{3. Portrait and Human-Centered Editing}\par\vspace{1pt}
\noindent\hspace*{0.3em}\textbf{Face Editing}\par
\noindent\hspace*{0.7em}\textbf{Beauty Makeup}\par
\noindent\hspace*{0.7em}\begin{tabular}{@{}p{0.45\linewidth}@{\hspace{0.02\linewidth}}p{0.45\linewidth}@{}}
$\bullet$ Auto Skin Smoothing & $\bullet$ Skin Whitening \\
$\bullet$ Acne \& Blemish Removal & $\bullet$ Remove Dark Circles \\
$\bullet$ Remove Nasolabial Folds & $\bullet$ Remove Tear Troughs \\
$\bullet$ Remove Neck Lines & $\bullet$ Remove Shine/Oiliness \\
$\bullet$ Pore Minimizer & $\bullet$ Skin Tone Temperature \\
$\bullet$ Tanning & $\bullet$ Slim Face \\
$\bullet$ Small Face & $\bullet$ Jawline Definition \\
$\bullet$ Cheekbone Reduction & $\bullet$ Temple Filling \\
$\bullet$ Chin Reshaping & $\bullet$ Forehead Height Adjustment \\
$\bullet$ Hairline Filling & $\bullet$ Enlarge Eyes \\
$\bullet$ Eye Brightening & $\bullet$ Eye Distance Adjustment \\
$\bullet$ Eye Tilt/Angle & $\bullet$ Double Eyelid Generation \\
$\bullet$ Aegyo-sal (Under-eye fat) & $\bullet$ Red Eye Removal \\
$\bullet$ Slim Nose & $\bullet$ Nose Bridge Lift \\
$\bullet$ Nostril Reduction & $\bullet$ Nose Tip Reshaping \\
$\bullet$ Philtrum Shortening & $\bullet$ Lip Plumping \\
$\bullet$ Smile Lift & $\bullet$ Teeth Whitening \\
$\bullet$ Teeth Correction & $\bullet$ Lip Shape Adjustment \\
$\bullet$ 3D Contouring & $\bullet$ Face Highlighting \\
$\bullet$ Eyebrow Reshaping & $\bullet$ Eyebrow Density \\
$\bullet$ Eyelash Extension & $\bullet$ Lower Eyelid Down \\
$\bullet$ Facial Asymmetry Correction &  \\
\end{tabular}\par\vspace{1pt}
\noindent\hspace*{0.7em}\textbf{Facial Attributes}\par
\noindent\hspace*{0.7em}\begin{tabular}{@{}p{0.45\linewidth}@{\hspace{0.02\linewidth}}p{0.45\linewidth}@{}}
$\bullet$ Make older & $\bullet$ Make younger \\
$\bullet$ Baby face & $\bullet$ Gender swap \\
$\bullet$ Add bangs & $\bullet$ Long hair \\
$\bullet$ Short hair & $\bullet$ Curly hair \\
$\bullet$ Straight hair & $\bullet$ Make bald \\
$\bullet$ Buzz cut & $\bullet$ Dreadlocks \\
$\bullet$ Twin tails & $\bullet$ Blonde hair \\
$\bullet$ Black hair & $\bullet$ Red hair \\
$\bullet$ Silver/White hair & $\bullet$ Brown hair \\
$\bullet$ Highlights/Ombre hair & $\bullet$ Add full beard \\
$\bullet$ Add mustache & $\bullet$ Add goatee \\
$\bullet$ Remove beard & $\bullet$ Add freckles \\
$\bullet$ Tanned skin & $\bullet$ Pale/Fair skin \\
$\bullet$ Change eye color & $\bullet$ Add eyeglasses \\
$\bullet$ Add sunglasses & $\bullet$ Remove glasses \\
$\bullet$ Add hat & $\bullet$ Add baseball cap \\
$\bullet$ Add earrings & $\bullet$ Add necklace \\
$\bullet$ Add face mask & $\bullet$ Heavy makeup \\
$\bullet$ Remove makeup & $\bullet$ Change lipstick color \\
$\bullet$ Double eyelids &  \\
\end{tabular}\par\vspace{1pt}
\noindent\hspace*{0.7em}\textbf{Hairstyle Editing}\par
\noindent\hspace*{0.7em}\begin{tabular}{@{}p{0.45\linewidth}@{\hspace{0.02\linewidth}}p{0.45\linewidth}@{}}
$\bullet$ Add Bangs & $\bullet$ French Bangs \\
$\bullet$ Curtain Bangs & $\bullet$ Straight Bangs \\
$\bullet$ Side-swept Bangs & $\bullet$ Baby Hair Bangs \\
$\bullet$ Long Hair & $\bullet$ Short Hair \\
$\bullet$ Curly Hair & $\bullet$ Big Wavy Hair \\
$\bullet$ Fleece Curls & $\bullet$ Straight Hair \\
$\bullet$ Smooth Hair & $\bullet$ Bald \\
$\bullet$ Buzz Cut & $\bullet$ Dreadlocks \\
$\bullet$ Twin Tails & $\bullet$ High Ponytail \\
$\bullet$ Hair Bun & $\bullet$ Bob Cut \\
$\bullet$ Slicked Back & $\bullet$ Middle Part \\
$\bullet$ Side Part & $\bullet$ Wolf Cut/Mullet \\
$\bullet$ Hime Cut & $\bullet$ Afro \\
$\bullet$ Braids & $\bullet$ Updo \\
$\bullet$ Undercut & $\bullet$ Fill Hairline \\
$\bullet$ Recede Hairline & $\bullet$ Increase Hair Volume \\
$\bullet$ Volumize Hair Roots & $\bullet$ Remove Flyaways \\
$\bullet$ Frizz Control & $\bullet$ Enhance Hair Shine \\
$\bullet$ Wet Hair Effect & $\bullet$ Remove Greasy Hair \\
$\bullet$ Hair Dye & $\bullet$ Blonde Hair \\
$\bullet$ Black Hair & $\bullet$ Red Hair \\
$\bullet$ Silver/White Hair & $\bullet$ Brown Hair \\
$\bullet$ Flaxen/Ash Brown & $\bullet$ Rose Gold \\
$\bullet$ Smoky Blue & $\bullet$ Hair Highlights \\
$\bullet$ Inner Hair Color & $\bullet$ Ombre Hair \\
$\bullet$ Ash Blonde & $\bullet$ Pink Hair \\
$\bullet$ Add Full Beard & $\bullet$ Add Mustache \\
$\bullet$ Add Goatee & $\bullet$ Remove Beard \\
$\bullet$ Stubble Effect & $\bullet$ Sideburns Trim \\
$\bullet$ Eyebrow Shaping & $\bullet$ Thicken Eyebrows \\
$\bullet$ Feathery Eyebrows & $\bullet$ Thin Eyebrows \\
$\bullet$ Straight Eyebrows & $\bullet$ Arched Eyebrows \\
\end{tabular}\par\vspace{1pt}
\noindent\hspace*{0.7em}\textbf{Emotion Expression}\par
\noindent\hspace*{0.7em}\begin{tabular}{@{}p{0.45\linewidth}@{\hspace{0.02\linewidth}}p{0.45\linewidth}@{}}
$\bullet$ Smile & $\bullet$ Laugh \\
$\bullet$ Closed-mouth Smile & $\bullet$ Smirk \\
$\bullet$ Bitter Smile & $\bullet$ Giggle \\
$\bullet$ Sadness & $\bullet$ Crying \\
$\bullet$ Anger & $\bullet$ Frown \\
$\bullet$ Surprise & $\bullet$ Fear \\
$\bullet$ Disgust & $\bullet$ Contempt \\
$\bullet$ Confusion & $\bullet$ Serious \\
$\bullet$ Poker Face/Cool & $\bullet$ Bored/Apathetic \\
$\bullet$ Pout & $\bullet$ Tongue Out \\
$\bullet$ Bite Lip & $\bullet$ Blow Kiss \\
$\bullet$ Wink & $\bullet$ Close Eyes \\
$\bullet$ Roll Eyes & $\bullet$ Open Mouth/Gasp \\
$\bullet$ Scream & $\bullet$ Yawn \\
$\bullet$ Shy/Blush & $\bullet$ Flirty/Seductive \\
$\bullet$ Confident & $\bullet$ Tired \\
$\bullet$ Tipsy/Drunk & $\bullet$ Pain \\
$\bullet$ Relieved & $\bullet$ Exaggerate Expression \\
$\bullet$ Subtle Expression &  \\
\end{tabular}\par\vspace{1pt}
\noindent\hspace*{0.7em}\textbf{Gaze Correction}\par
\noindent\hspace*{0.7em}\begin{tabular}{@{}p{0.45\linewidth}@{\hspace{0.02\linewidth}}p{0.45\linewidth}@{}}
$\bullet$ Look at Camera & $\bullet$ Adjust Gaze Direction \\
$\bullet$ Open Closed Eyes & $\bullet$ Add Eye Catchlights \\
$\bullet$ Remove Red-eye & $\bullet$ Fix Cross-eyed \\
$\bullet$ Strabismus Correction & $\bullet$ Whiten Sclera \\
$\bullet$ Balance Asymmetric Eyes & $\bullet$ Enhance Iris Texture \\
$\bullet$ Fix Lifeless Eyes & $\bullet$ Generate Wink \\
$\bullet$ Simulate Squint & $\bullet$ Lift Droopy Eyelids \\
$\bullet$ Focus Gaze & $\bullet$ Soften Gaze \\
$\bullet$ Sharpen Gaze & $\bullet$ Adjust Eye Distance \\
$\bullet$ Resize Pupil & $\bullet$ Watery Eyes Effect \\
\end{tabular}\par\vspace{1pt}
\noindent\hspace*{0.3em}\textbf{Body and Fashion}\par
\noindent\hspace*{0.7em}\textbf{Virtual Try On}\par
\noindent\hspace*{0.7em}\begin{tabular}{@{}p{0.45\linewidth}@{\hspace{0.02\linewidth}}p{0.45\linewidth}@{}}
$\bullet$ Change Top & $\bullet$ Change Bottoms \\
$\bullet$ Full Outfit Change & $\bullet$ Try on Dress \\
$\bullet$ Try on Hoodie & $\bullet$ Try on Shirt \\
$\bullet$ Try on Jeans & $\bullet$ Try on Skirt \\
$\bullet$ Try on Suit & $\bullet$ Try on Wedding Dress \\
$\bullet$ Try on Hanfu/Costume & $\bullet$ Try on Swimwear \\
$\bullet$ Try on Sportswear & $\bullet$ Try on Coat/Trench \\
$\bullet$ Try on Puffer Jacket & $\bullet$ Flat Lay to Model \\
$\bullet$ Mannequin to Model & $\bullet$ Ghost Mannequin Effect \\
$\bullet$ Preserve Logo Details & $\bullet$ Maintain Fabric Texture \\
$\bullet$ Recolor Garment & $\bullet$ Replace Clothing Pattern \\
$\bullet$ Adjust Hemline & $\bullet$ Oversized Fit \\
$\bullet$ Slim Fit & $\bullet$ Tuck in Shirt \\
$\bullet$ Untucked Shirt & $\bullet$ Open Jacket \\
$\bullet$ Zip Up & $\bullet$ Roll up Sleeves \\
$\bullet$ Try on Glasses & $\bullet$ Try on Hat \\
$\bullet$ Try on Jewelry/Necklace & $\bullet$ Try on Shoes \\
$\bullet$ Try on Handbag & $\bullet$ Generate Virtual Model \\
$\bullet$ Batch E-commerce Try-on & $\bullet$ Street Snap Style \\
$\bullet$ Studio Lighting Style & $\bullet$ Lingerie Model Gen \\
$\bullet$ Change Outfit Style & $\bullet$ Fix Garment Distortion \\
\end{tabular}\par\vspace{1pt}
\noindent\hspace*{0.7em}\textbf{Body Reshape}\par
\noindent\hspace*{0.7em}\begin{tabular}{@{}p{0.45\linewidth}@{\hspace{0.02\linewidth}}p{0.45\linewidth}@{}}
$\bullet$ Auto Body Slimming & $\bullet$ Leg Lengthening \\
$\bullet$ Waist Slimming & $\bullet$ Arm Slimming \\
$\bullet$ Hip Enhancement & $\bullet$ Head Size Reduction \\
$\bullet$ Swan Neck & $\bullet$ Shoulder Width Adjustment \\
$\bullet$ Right-angled Shoulders & $\bullet$ Breast Enhancement \\
$\bullet$ Abs Definition & $\bullet$ Muscle Line Enhancement \\
$\bullet$ Thigh Slimming & $\bullet$ Calf Slimming \\
$\bullet$ Full Body Height & $\bullet$ Posture Correction \\
$\bullet$ Hunchback Correction & $\bullet$ Collarbone Definition \\
$\bullet$ Body Proportion Adjustment & $\bullet$ Flatten Belly \\
\end{tabular}\par\vspace{1pt}
\noindent\hspace*{0.7em}\textbf{Pose Action}\par
\noindent\hspace*{0.7em}\begin{tabular}{@{}p{0.45\linewidth}@{\hspace{0.02\linewidth}}p{0.45\linewidth}@{}}
$\bullet$ Reference pose transfer & $\bullet$ Custom skeleton \\
$\bullet$ Turn head & $\bullet$ Look up \\
$\bullet$ Look down & $\bullet$ Tilt head \\
$\bullet$ Look at camera & $\bullet$ Look away \\
$\bullet$ Fix deformed hands & $\bullet$ Refine fingers \\
$\bullet$ Peace sign & $\bullet$ Finger heart \\
$\bullet$ Thumbs up & $\bullet$ Waving \\
$\bullet$ Pointing & $\bullet$ Clenched fist \\
$\bullet$ Open palm & $\bullet$ Praying hands \\
$\bullet$ Arms crossed & $\bullet$ Hands on hips \\
$\bullet$ Hands in pockets & $\bullet$ Hands behind head \\
$\bullet$ Raise hands & $\bullet$ Stretching \\
$\bullet$ Standing straight & $\bullet$ Sitting \\
$\bullet$ Sitting cross-legged & $\bullet$ Cross legs \\
$\bullet$ Squatting & $\bullet$ Kneeling \\
$\bullet$ Lying down & $\bullet$ Lying on side \\
$\bullet$ Leaning & $\bullet$ Walking \\
$\bullet$ Running & $\bullet$ Jumping \\
$\bullet$ Dancing & $\bullet$ Yoga pose \\
$\bullet$ Kicking & $\bullet$ Turn around \\
$\bullet$ Side profile & $\bullet$ Holding phone \\
$\bullet$ Holding cup & $\bullet$ Hugging \\
\end{tabular}\par\vspace{1pt}
\vspace{3pt}\noindent\rule{\linewidth}{0.4pt}\par
\noindent\textbf{4. Text and Graphic Design}\par\vspace{1pt}
\noindent\hspace*{0.3em}\textbf{Text Manipulation}\par
\noindent\hspace*{0.7em}\textbf{Text Removal}\par
\noindent\hspace*{0.7em}\begin{tabular}{@{}p{0.45\linewidth}@{\hspace{0.02\linewidth}}p{0.45\linewidth}@{}}
$\bullet$ Smart Text Removal & $\bullet$ Remove Watermark \\
$\bullet$ Remove TV Logo & $\bullet$ Remove Subtitles \\
$\bullet$ Remove Date Stamp & $\bullet$ Remove Handwriting \\
$\bullet$ Remove Stamp/Seal & $\bullet$ Remove Manga Text \\
$\bullet$ Remove Street Sign Text & $\bullet$ Remove License Plate \\
$\bullet$ Remove Text on Clothing & $\bullet$ Remove Product Logo \\
$\bullet$ Remove Screenshot UI & $\bullet$ Remove Bullet Comments \\
$\bullet$ Remove Copyright Symbol & $\bullet$ Remove Highlighter \\
$\bullet$ Clear Exam Answers & $\bullet$ Remove Camera Watermark \\
$\bullet$ Remove Graffiti & $\bullet$ Remove Text from Complex Background \\
\end{tabular}\par\vspace{1pt}
\noindent\hspace*{0.7em}\textbf{Text Editing}\par
\noindent\hspace*{0.7em}\begin{tabular}{@{}p{0.45\linewidth}@{\hspace{0.02\linewidth}}p{0.45\linewidth}@{}}
$\bullet$ Scene Text Modification & $\bullet$ Translate Text in Image \\
$\bullet$ Handwriting Generation & $\bullet$ Calligraphy Style \\
$\bullet$ 3D Text Effect & $\bullet$ Neon Sign Text \\
$\bullet$ Graffiti Text & $\bullet$ Elemental Text Effects (Fire/Water) \\
$\bullet$ Metallic/Gold Foil Text & $\bullet$ Engraved \& Embossed Text \\
$\bullet$ Embroidery Style Text & $\bullet$ Chalk/Crayon Text \\
$\bullet$ Perspective Text Matching & $\bullet$ Curved Surface Text \\
$\bullet$ Artistic Font Generation & $\bullet$ Meme Captioning \\
$\bullet$ Retro Pixel Text & $\bullet$ Text Texture Replacement \\
$\bullet$ Poster Headline Design & $\bullet$ Glowing Text Effect \\
\end{tabular}\par\vspace{1pt}
\noindent\hspace*{0.3em}\textbf{Design Elements}\par
\noindent\hspace*{0.7em}\textbf{Typography Style}\par
\noindent\hspace*{0.7em}\begin{tabular}{@{}p{0.45\linewidth}@{\hspace{0.02\linewidth}}p{0.45\linewidth}@{}}
$\bullet$ 3D Text & $\bullet$ Neon Text \\
$\bullet$ Metallic Text & $\bullet$ Handwritten Style \\
$\bullet$ Calligraphy & $\bullet$ Graffiti Style \\
$\bullet$ Pixel Text & $\bullet$ Glitch Text \\
$\bullet$ Fire Effect & $\bullet$ Ice Effect \\
$\bullet$ Bubble Text & $\bullet$ Gothic Style \\
$\bullet$ Retro Serif & $\bullet$ Golden Text \\
$\bullet$ Liquid Text & $\bullet$ Glass Texture \\
$\bullet$ Stone Carving & $\bullet$ Wood Texture \\
$\bullet$ Chalk Style & $\bullet$ Ink Style \\
$\bullet$ Cyberpunk Style & $\bullet$ Balloon Text \\
$\bullet$ Furry Text & $\bullet$ Gradient Text \\
$\bullet$ Outline Text & $\bullet$ Shadow Text \\
$\bullet$ Sticker Style & $\bullet$ Floral Text \\
$\bullet$ Food Texture & $\bullet$ Glowing Text \\
\end{tabular}\par\vspace{1pt}
\noindent\hspace*{0.7em}\textbf{Layout Logo}\par
\noindent\hspace*{0.7em}\begin{tabular}{@{}p{0.45\linewidth}@{\hspace{0.02\linewidth}}p{0.45\linewidth}@{}}
$\bullet$ Poster Layout & $\bullet$ Smart Composition \\
$\bullet$ Photo Collage & $\bullet$ Magazine Cover Design \\
$\bullet$ E-commerce Page Layout & $\bullet$ Social Media Templates \\
$\bullet$ Logo Generation & $\bullet$ Insert Logo \\
$\bullet$ Make Logo Transparent & $\bullet$ Logo Vectorization \\
$\bullet$ Logo Stylization & $\bullet$ 3D Logo Effect \\
$\bullet$ Tiled Watermark & $\bullet$ Invisible Watermark \\
$\bullet$ Insert QR Code & $\bullet$ Artistic QR Code \\
$\bullet$ Add Borders & $\bullet$ Add Stickers \\
$\bullet$ Auto Alignment & $\bullet$ Negative Space Management \\
\end{tabular}\par\vspace{1pt}
\vspace{3pt}\noindent\rule{\linewidth}{0.4pt}\par
\noindent\textbf{5. Generation and Composition}\par\vspace{1pt}
\noindent\hspace*{0.3em}\textbf{Canvas and Viewpoint}\par
\noindent\hspace*{0.7em}\textbf{Outpainting}\par
\noindent\hspace*{0.7em}\begin{tabular}{@{}p{0.45\linewidth}@{\hspace{0.02\linewidth}}p{0.45\linewidth}@{}}
$\bullet$ Horizontal Expansion & $\bullet$ Vertical Expansion \\
$\bullet$ Expand All Sides & $\bullet$ Smart Autofill \\
$\bullet$ Subject Re-centering & $\bullet$ Fit to Wallpaper \\
$\bullet$ Background Extension & $\bullet$ Complete Cut-off Objects \\
$\bullet$ Panorama Generation & $\bullet$ Fill Rotated Corners \\
$\bullet$ 1:1 Square Expansion & $\bullet$ Feathered Expansion \\
$\bullet$ Zoom Out (Uncrop) &  \\
\end{tabular}\par\vspace{1pt}
\noindent\hspace*{0.7em}\textbf{Crop Composition}\par
\noindent\hspace*{0.7em}\begin{tabular}{@{}p{0.45\linewidth}@{\hspace{0.02\linewidth}}p{0.45\linewidth}@{}}
$\bullet$ ID Photo Crop & $\bullet$ Smart Subject Centering \\
$\bullet$ Auto Straighten & $\bullet$ Perspective Crop \\
$\bullet$ Rule of Thirds & $\bullet$ Golden Ratio \\
$\bullet$ Cinematic Aspect Ratio &  \\
\end{tabular}\par\vspace{1pt}
\noindent\hspace*{0.7em}\textbf{Camera Shift}\par
\noindent\hspace*{0.7em}\begin{tabular}{@{}p{0.45\linewidth}@{\hspace{0.02\linewidth}}p{0.45\linewidth}@{}}
$\bullet$ Front View & $\bullet$ Side View \\
$\bullet$ Back View & $\bullet$ High Angle / Bird's Eye \\
$\bullet$ Low Angle / Worm's Eye & $\bullet$ Three-quarter View \\
$\bullet$ First-Person View & $\bullet$ Selfie Angle \\
$\bullet$ Over-the-Shoulder & $\bullet$ Drone Shot \\
$\bullet$ Isometric View & $\bullet$ Wide Angle \\
$\bullet$ Fisheye Lens & $\bullet$ Macro / Close-up \\
$\bullet$ Panoramic View & $\bullet$ Zoom In \\
$\bullet$ Zoom Out & $\bullet$ Camera Pan \\
$\bullet$ Perspective Correction & $\bullet$ 3D Rotation \\
\end{tabular}\par\vspace{1pt}
\noindent\hspace*{0.3em}\textbf{Conditional Generation}\par
\noindent\hspace*{0.7em}\textbf{Reference Driven}\par
\noindent\hspace*{0.7em}\begin{tabular}{@{}p{0.45\linewidth}@{\hspace{0.02\linewidth}}p{0.45\linewidth}@{}}
$\bullet$ Style Transfer & $\bullet$ Color Palette Matching \\
$\bullet$ Composition Reference & $\bullet$ Human Pose Copy \\
$\bullet$ Face Identity Lock & $\bullet$ Character IP Consistency \\
$\bullet$ Line Art Colorization & $\bullet$ Spatial Structure Reference \\
$\bullet$ Depth Reference & $\bullet$ Edge Outline Lock \\
$\bullet$ Sketch to Realistic & $\bullet$ Anime to Photorealistic \\
$\bullet$ Photorealistic to Anime & $\bullet$ Generate Variations \\
$\bullet$ Outfit Style Reference & $\bullet$ Hairstyle Reference \\
$\bullet$ Makeup Reference & $\bullet$ Material Texture Copy \\
$\bullet$ Lighting Layout Reference & $\bullet$ Atmosphere Copy \\
$\bullet$ Background Reference & $\bullet$ Product Design Reference \\
$\bullet$ Interior Design Reference & $\bullet$ Architectural Structure Reference \\
$\bullet$ Motion Capture & $\bullet$ Expression Transfer \\
$\bullet$ Hand Gesture Reference & $\bullet$ Logo Shape Reference \\
$\bullet$ Artistic Brushstroke Copy & $\bullet$ Local Area Reference \\
$\bullet$ Semantic Segmentation Guide & $\bullet$ Artistic QR Code \\
$\bullet$ Film Aesthetic Copy & $\bullet$ 3D Render Reference \\
\end{tabular}\par\vspace{1pt}
\noindent\hspace*{0.7em}\textbf{Sketch Control}\par
\noindent\hspace*{0.7em}\begin{tabular}{@{}p{0.45\linewidth}@{\hspace{0.02\linewidth}}p{0.45\linewidth}@{}}
$\bullet$ Sketch to Realistic Photo & $\bullet$ Scribble to Art \\
$\bullet$ Line Art Colorization & $\bullet$ Manga/Anime Coloring \\
$\bullet$ Architectural Sketch Rendering & $\bullet$ Interior Design Rendering \\
$\bullet$ Product Design Rendering & $\bullet$ Fashion Sketch Rendering \\
$\bullet$ Refine Rough Sketch & $\bullet$ Silhouette to Image \\
$\bullet$ Generative Fill & $\bullet$ Local Redraw \\
$\bullet$ Add Element via Brush & $\bullet$ Texture Inpainting \\
$\bullet$ Fix Hands \& Limbs & $\bullet$ Face Inpainting \\
$\bullet$ Color Block Composition & $\bullet$ Palette Guided Generation \\
$\bullet$ Structure-Preserved Redraw &  \\
\end{tabular}\par\vspace{1pt}
\noindent\hspace*{0.7em}\textbf{Multi Image Consistency}\par
\noindent\hspace*{0.7em}\begin{tabular}{@{}p{0.45\linewidth}@{\hspace{0.02\linewidth}}p{0.45\linewidth}@{}}
$\bullet$ Double Exposure & $\bullet$ Image Blending \\
$\bullet$ Smart Collage & $\bullet$ Long Image Stitching \\
$\bullet$ Panorama Stitching & $\bullet$ 360 Panorama \\
$\bullet$ Face Swap & $\bullet$ Head Swap \\
$\bullet$ Character Consistency & $\bullet$ Keep Character Change Pose \\
$\bullet$ Keep Character Change Background & $\bullet$ Outfit Consistency \\
$\bullet$ Multi-Angle Generation & $\bullet$ 3-View Generation \\
$\bullet$ Character Sheet & $\bullet$ Comic Panel Generation \\
$\bullet$ 4-Panel Comic & $\bullet$ Picture Book Consistency \\
$\bullet$ Storyboard Generation & $\bullet$ Cinematic Storyboard \\
$\bullet$ Image Morphing & $\bullet$ Style Mixing \\
$\bullet$ Concept Blending & $\bullet$ Seamless Texture Tiling \\
$\bullet$ HDR Merge & $\bullet$ Focus Stacking \\
$\bullet$ Creative Compositing & $\bullet$ Montage Effect \\
$\bullet$ Collage Art & $\bullet$ Photo Bash \\
\end{tabular}\par\vspace{1pt}
\noindent\hspace*{0.7em}\textbf{Group Photo Synthesis}\par
\noindent\hspace*{0.7em}\begin{tabular}{@{}p{0.45\linewidth}@{\hspace{0.02\linewidth}}p{0.45\linewidth}@{}}
$\bullet$ Celebrity Group Photo & $\bullet$ Virtual Idol Co-framing \\
$\bullet$ Add Person to Group Photo & $\bullet$ Remove Person from Group Photo \\
$\bullet$ Multi-person Face Swap & $\bullet$ Multi-person Outfit Swap \\
$\bullet$ Family Portrait Generation & $\bullet$ BFF Photo Synthesis \\
$\bullet$ Couple Photo Synthesis & $\bullet$ Wedding Photo Synthesis \\
$\bullet$ Cross-time Group Photo & $\bullet$ Swap Character Positions \\
$\bullet$ Center Position Adjustment & $\bullet$ Height Proportion Adjustment \\
$\bullet$ Multi-person Lighting Unification & $\bullet$ Multi-person Color Tone Matching \\
$\bullet$ Multi-person Perspective Correction & $\bullet$ Unified Skin Texture \\
$\bullet$ Fix Closed Eyes in Group Photo & $\bullet$ Multi-person Expression Sync \\
$\bullet$ Eye Contact Alignment & $\bullet$ Hold Hands Generation \\
$\bullet$ Hugging Pose Generation & $\bullet$ Put Arm Around Shoulder \\
$\bullet$ Back-to-Back Pose & $\bullet$ Multi-character Consistency \\
$\bullet$ Dense Crowd Generation & $\bullet$ Party Scene Generation \\
$\bullet$ Business Meeting Group Photo & $\bullet$ Graduation Photo Generation \\
$\bullet$ Team Uniform Unification & $\bullet$ Remove Passersby from Background \\
$\bullet$ Fix Group Photo Edge Distortion & $\bullet$ Anime Character Crossover \\
$\bullet$ Cinematic Ensemble Poster & $\bullet$ Historical Figure Photo Replica \\
\end{tabular}\par\vspace{1pt}
\vspace{3pt}\noindent\rule{\linewidth}{0.4pt}\par
\noindent\textbf{6. Advanced and Domain-Specific Applications}\par\vspace{1pt}
\noindent\hspace*{0.3em}\textbf{Reasoning and Interaction}\par
\noindent\hspace*{0.7em}\textbf{Complex Instruction}\par
\noindent\hspace*{0.7em}\begin{tabular}{@{}p{0.45\linewidth}@{\hspace{0.02\linewidth}}p{0.45\linewidth}@{}}
$\bullet$ Multiple Condition Stacking & $\bullet$ Negative Constraints Handling \\
$\bullet$ Precise Quantity Control & $\bullet$ Spatial Position Specification \\
$\bullet$ Independent Multi-object Attributes & $\bullet$ Referential Disambiguation \\
$\bullet$ Logical Causal Inference & $\bullet$ Comparative Instructions \\
$\bullet$ Abstract Concept Visualization & $\bullet$ Sequential Multi-step Operations \\
$\bullet$ Modification with Preservation & $\bullet$ Physics Common Sense Adherence \\
$\bullet$ Implicit Intent Inference & $\bullet$ Counterfactual Editing \\
$\bullet$ Complex Composition Description & $\bullet$ Relative Size Adjustment \\
$\bullet$ Specific Style Fusion & $\bullet$ Reference-based Modification \\
$\bullet$ Exclusionary Editing & $\bullet$ Multi-level Detail Description \\
\end{tabular}\par\vspace{1pt}
\noindent\hspace*{0.7em}\textbf{Logic Process}\par
\noindent\hspace*{0.7em}\begin{tabular}{@{}p{0.45\linewidth}@{\hspace{0.02\linewidth}}p{0.45\linewidth}@{}}
$\bullet$ Storyboard Generation & $\bullet$ Evolutionary Process \\
$\bullet$ Assembly Instructions & $\bullet$ Cooking Steps Breakdown \\
$\bullet$ Science Experiment Procedure & $\bullet$ Exact Count Generation \\
$\bullet$ Spatial Position Constraints & $\bullet$ Floor Plan Generation \\
$\bullet$ Component Breakdown (Knolling) & $\bullet$ Cross-Section View \\
$\bullet$ Before and After Comparison & $\bullet$ Cause and Effect Visualization \\
$\bullet$ Anatomical Structure & $\bullet$ Infographic Generation \\
$\bullet$ Historical Timeline & $\bullet$ Spot the Difference Game \\
$\bullet$ Maze Generation & $\bullet$ Optical Illusion \\
$\bullet$ Hidden Object Game & $\bullet$ Calligram (Text as Objects) \\
$\bullet$ Visual Pun & $\bullet$ Physics Simulation \\
$\bullet$ Logic Puzzle Illustration & $\bullet$ Flowchart Visualization \\
$\bullet$ Comparison Diagram & $\bullet$ Multi-View Orthographic \\
$\bullet$ Inclusion Relationship & $\bullet$ Exclusion Logic Diagram \\
$\bullet$ Cyclic Process Diagram & $\bullet$ Hierarchy Diagram \\
\end{tabular}\par\vspace{1pt}
\noindent\hspace*{0.3em}\textbf{Industry Solutions}\par
\noindent\hspace*{0.7em}\textbf{Ecommerce}\par
\noindent\hspace*{0.7em}\begin{tabular}{@{}p{0.45\linewidth}@{\hspace{0.02\linewidth}}p{0.45\linewidth}@{}}
$\bullet$ AI Commercial Photography & $\bullet$ Product Background Replacement \\
$\bullet$ Mannequin to Model & $\bullet$ Virtual Try-on \\
$\bullet$ Model Face Swap/Localization & $\bullet$ Product Color Variation (SKU) \\
$\bullet$ AI Podium \& Stand Generation & $\bullet$ Product Shadow \& Reflection \\
$\bullet$ White Background Shot & $\bullet$ E-commerce Poster Generation \\
$\bullet$ Promotional Banner Design & $\bullet$ Lifestyle Context Integration \\
$\bullet$ Product Infographic & $\bullet$ Packaging Design Mockup \\
$\bullet$ Apparel Pattern Preview & $\bullet$ Social Media Marketing Image \\
$\bullet$ Amazon Main Image Optimization & $\bullet$ Food Photography Enhancement \\
$\bullet$ Jewelry Sparkle Effect & $\bullet$ Furniture Scene Staging \\
$\bullet$ Seasonal Marketing Theme & $\bullet$ Batch Product Cutout \\
$\bullet$ Listing Long Image Stitching & $\bullet$ Storefront Decoration Assets \\
$\bullet$ Brand Logo Integration & $\bullet$ 3D Product Rendering \\
$\bullet$ Model Age/Ethnicity Adjustment & $\bullet$ Material Detail Zoom \\
$\bullet$ Cosmetic Texture Display & $\bullet$ Flat Lay Generation \\
$\bullet$ Digital Screen Replacement & $\bullet$ Outfit Style Recommendation \\
\end{tabular}\par\vspace{1pt}
\noindent\hspace*{0.7em}\textbf{Docs Education}\par
\noindent\hspace*{0.7em}\begin{tabular}{@{}p{0.45\linewidth}@{\hspace{0.02\linewidth}}p{0.45\linewidth}@{}}
$\bullet$ Remove Document Shadows & $\bullet$ Document Straightening \& Dewarping \\
$\bullet$ Document Background Whitening & $\bullet$ Remove Screen Moire Patterns \\
$\bullet$ Book Page Scanning Enhancement & $\bullet$ Remove Page Creases \\
$\bullet$ Old Document Restoration & $\bullet$ Blueprint \& Technical Drawing Enhancement \\
$\bullet$ Receipt Clarification & $\bullet$ Remove Handwriting from Exam \\
$\bullet$ Remove Grading Marks & $\bullet$ Blackboard Writing Clarification \\
$\bullet$ Whiteboard Glare Removal & $\bullet$ Error Problem Notebook Gen \\
$\bullet$ Math Formula Beautification & $\bullet$ Flashcard Generation \\
$\bullet$ ID Photo Generation & $\bullet$ ID Photo Background Change \\
$\bullet$ ID Photo Smart Layout & $\bullet$ ID Photo Virtual Suit Try-on \\
$\bullet$ Resume Photo Retouching & $\bullet$ ID Card Copy Generation \\
$\bullet$ Digital Signature Extraction & $\bullet$ Stamp \& Seal Extraction \\
$\bullet$ Table Structure Restoration & $\bullet$ Business Card Digitization \\
$\bullet$ PPT Slide Generation & $\bullet$ UI Interface Generation \\
$\bullet$ Sketch to UI Design & $\bullet$ Mind Map Generation \\
$\bullet$ Mind Map Beautification & $\bullet$ Journal Sticker Generation \\
$\bullet$ Note Layout Optimization & $\bullet$ Data Chart Generation \\
$\bullet$ Educational Illustration Gen & $\bullet$ Certificate \& Award Creation \\
$\bullet$ Infographic Design & $\bullet$ Handwriting Simulation \\
\end{tabular}\par\vspace{1pt}
\noindent\hspace*{0.7em}\textbf{Entertainment Acg}\par
\noindent\hspace*{0.7em}\begin{tabular}{@{}p{0.45\linewidth}@{\hspace{0.02\linewidth}}p{0.45\linewidth}@{}}
$\bullet$ Character Design Sheet & $\bullet$ Character Turnaround \\
$\bullet$ Chibi/Nendoroid Style & $\bullet$ Pixel Art \\
$\bullet$ Game Icons \& Props & $\bullet$ Concept Art \\
$\bullet$ Manga Page/Layout & $\bullet$ Webtoon/Manhwa \\
$\bullet$ Visual Novel Backgrounds & $\bullet$ TCG Card Illustration \\
$\bullet$ 90s Retro Anime & $\bullet$ Mecha \& Robot Design \\
$\bullet$ Monster \& Creature Design & $\bullet$ Weapon \& Equipment Design \\
$\bullet$ Game Sprite Sheets & $\bullet$ Isometric/2.5D View \\
$\bullet$ Low Poly Style & $\bullet$ Voxel Art \\
$\bullet$ Impasto/Thick Painting & $\bullet$ Cell Shading \\
$\bullet$ Light Novel Cover & $\bullet$ Sticker/Emoji Generation \\
$\bullet$ Furry/Anthro & $\bullet$ Game UI Assets \\
$\bullet$ Magic \& Skill VFX & $\bullet$ Seamless Textures \\
$\bullet$ Storyboard & $\bullet$ Character Variations \\
$\bullet$ Makoto Shinkai Style & $\bullet$ Cyberpunk Anime \\
$\bullet$ Vtuber Avatar & $\bullet$ Fan Art Illustration \\
$\bullet$ Game Map Tiles & $\bullet$ Ghibli Style \\
$\bullet$ American Comic Style & $\bullet$ Figurine/Model Rendering \\
\end{tabular}\par\vspace{1pt}
\noindent\rule{\linewidth}{0.4pt}\par
\endgroup

\end{document}